%% file: glrm_paper_supplementary.tex
\documentclass{article}

 \PassOptionsToPackage{numbers, compress}{natbib}

\usepackage[final]{neurips_2018}

\usepackage[colorlinks=true,
linkcolor=red,
urlcolor=blue,
citecolor=blue]{hyperref}

\usepackage[utf8]{inputenc} 
\usepackage[T1]{fontenc}    
\usepackage{url}            
\usepackage{booktabs}       
\usepackage{amsfonts}       
\usepackage{nicefrac}       
\usepackage{microtype}      
\usepackage{shortcuts_OPT}
\usepackage{bm,amsmath,multicol,caption}
\usepackage{amssymb}
\usepackage{algorithm,algorithmic}
\usepackage{xargs}
\usepackage{cleveref}
\usepackage{graphicx}
\usepackage{longtable}
\usepackage{answers}
\usepackage{array}
\usepackage{enumitem}
\usepackage{subcaption}

\newtheorem{Lemma}{Lemma}

\newtheorem{Theorem}{Theorem}

\newtheorem{assumption}{H\!}

\newcommand{\xupdownarrow}[1]{%
  {\left\updownarrow\vbox to #1{}\right.\kern-\nulldelimiterspace}
}

\def\rset{\mathbb{R}}

\def\Yset{\mathsf{Y}}

\def\Xset{\mathsf{X}}

\newcommand{\TX}{\mathbf{M}^0}
\newcommand{\Talpha}{\alp}
\newcommand{\TL}{\Prm^0}
\newcommandx{\smax}{\sigma_{+}}
\newcommandx{\smin}{\sigma_{-}}
\newcommandx{\nint}[2]{[#2]}
\newcommandx{\fu}[1]{\mathsf{f}_U(#1)}
\newcommandx{\R}{\mathbb{R}}
\newcommandx{\prob}[1]{\mathbb{P}\left(#1 \right) }
\newcommandx{\e}{\mathrm{e}}

\newcommand{\diff}[1]{\nabla\mathcal{L}(#1)}
\newcommand{\umax}{d_{\mathbf{X}}}
\newcommandx{\lone}{\|\cdot \|_{1}}
\newcommandx{\linf}{\| \cdot \|_{\infty}}
\newcommandx{\lop}{\| \cdot \|}
\newcommandx{\lnuc}{\| \cdot \|_{*}}
\newcommandx{\pe}[1]{\mathbb{E}\left[#1 \right] }

\usepackage[textsize=footnotesize]{todonotes} 

\title{Low-rank Interaction with Sparse Additive Effects Model for Large Data Frames}

\author{
  Genevi\`{e}ve Robin\\
  Centre de Math\'ematiques Appliqu\'ees\\
  \'Ecole Polytechnique, XPOP, INRIA\\
  91120 Palaiseau, France \\
  \texttt{genevieve.robin@polytechnique.edu} \\
  \And
  Hoi-To Wai \\
  Department of SE\&EM \\
  The Chinese University of Hong Kong\\
  Shatin, Hong Kong \\
 \texttt{htwai@se.cuhk.edu.hk} \\
   \AND
   Julie Josse \\
  Centre de Math\'ematiques Appliqu\'ees\\
  \'Ecole Polytechnique, XPOP, INRIA\\
  91120 Palaiseau, France \\
   \texttt{julie.josse@polytechnique.edu} \\
   \And
   Olga Klopp \\
   ESSEC Business School \\
   CREST, ENSAE \\
   95021 Cergy, France\\
   \texttt{klopp@essec.edu} \\
   \And
   \'Eric Moulines \\
  Centre de Math\'ematiques Appliqu\'ees\\
  \'Ecole Polytechnique, XPOP, INRIA\\
  91120 Palaiseau, France \\
   \texttt{eric.moulines@polytechnique.edu} \\
}

\begin{document}

\maketitle

\begin{abstract}
Many applications of machine learning involve the analysis of large data frames -- matrices collecting heterogeneous measurements (binary, numerical, counts, etc.) across samples -- with missing values. Low-rank models, as studied by \citet{udell2014generalized}, are popular in this framework for tasks such as visualization, clustering and missing value imputation. Yet, available methods with statistical guarantees and efficient optimization do not allow explicit modeling of main additive effects such as row and column, or covariate effects. In this paper, we introduce a \textit{low-rank interaction and sparse additive effects} (LORIS) model which combines matrix regression on a dictionary and low-rank design, to estimate main effects and interactions simultaneously. We provide statistical guarantees in the form of upper bounds on the estimation error of both components. Then, we introduce a \textit{mixed coordinate gradient descent} (MCGD) method which provably converges sub-linearly to an optimal solution and is computationally efficient for large scale data sets. We show on simulated and survey data that the method has a clear advantage over current practices, which consist in dealing separately with additive effects in a preprocessing step.
\end{abstract}

\input{intro.tex}

\input{problem.tex}


\input{algorithm.tex}

\input{sim.tex}

\section{Acknowledgement}
The authors would like to thank for the useful comments from three anonymous reviewers.
HTW's work was supported by the grant NSF CCF-BSF 1714672.

\medskip

{
\small
\bibliographystyle{abbrvnat}
\bibliography{references}
}

\input{append_stats_guarantees.tex}

\input{append.tex}

\end{document}

%% file: intro.tex

\section{Introduction}
Recently, a lot of effort has been devoted towards the efficient analysis
of large data frames, a term coined by \citet{udell2014generalized}.
A data frame is a large table of heterogeneous data (binary, numerical, counts)  with missing entries,
where each row
represents an example and each column a feature.
In order to analyze them, a powerful technique is to use \emph{low-rank models}
that embed rows and columns of data frames into low-dimensional spaces
\citep{Kiers1991,pages2014multiple,udell2014generalized},
enabling effective data analytics such as clustering,
visualization and missing value imputation; see also \citep{review-low-rank} and the references therein.


Characterizing additive effects 
of side information -- such as covariates, row or column effects --
\emph{simultaneously} with low rank interactions is an important
extension to plain low-rank models.
For example, in data frames obtained from recommender systems,
user information and item characteristics are known to
influence the ratings in addition to interactions between users and items \citep{feuerverger2012}.
These modifications to the low rank model have been advocated in the 
statistics literature, but they have been implemented only for small data
frames \citep{Agresti13}. 


In the large-scale low-rank matrix estimation literature,
available methods either
do not take additive effects into account \citep{deLeeuw:2006:PCA, Landgraf15, udell2014generalized,liu2016pca, fithian2013scalable},
or only handle the numerical data \citep{softImpute,pmlr-v51-gu16}.
As a common heuristics for preprocessing, 
prior work such as \citep{Landgraf15,udell2014generalized} 
remove the
row and column means and apply some normalization of 
the row and column variance. 
We show in numerical experiments this apparently benign operation
is not appropriate for large and heterogenous data frames,
and can cause severe impairments in the analysis.

The present work investigates a generalization of previous contributions
in the analysis of data frames.
Our contributions can be summarized as follows.
\paragraph{Contributions}
We present a new framework
that is \emph{statistically} and
\emph{computationally} efficient
for analyzing large and incomplete heterogeneous  data frames.
\begin{itemize}
\item We describe in \Cref{problem} the \emph{low-rank interaction with sparse additive effects} (LORIS) model, which combines matrix regression
on a dictionary with low rank approximation.
We propose a convex doubly penalized quasi-maximum likelihood approach, where the rank constraint is relaxed with a nuclear norm penalty, to estimate the regression
coefficients and the low rank component simultaneously.
We establish non-asymptotic upper bounds on the estimation errors.
\item We propose in \Cref{algo} a Mixed Coordinate Gradient Descent (MCGD) method to solve  efficiently the LORIS estimation problem. It uses a mixed update strategy including a proximal update for the sparse component and a conditional gradient (CG) for the low-rank component. We show that the MCGD method converges to an $\epsilon$-optimal solution in $\mathcal{O}(1 / \epsilon)$ iterations. We also outline an extension to efficient distributed implementation.
\item We demonstrate in \Cref{simus} the efficacy of our method both in terms of estimation and imputation quality on simulated and survey data examples.
\end{itemize}
\paragraph{Related work}
Our statistical model and analysis are related to prior work on \textit{low-rank plus sparse matrix decomposition} \citep{Xu:2010:RPV,Candes:2011:RPC,chandra,Hsu_robustmatrix,Klopp2017}; these papers provide statistical results for a particular case where the loss function is quadratic and the sparse component is entry-wise sparse. In comparison, the originality of the present work is two-fold. First, the sparsity pattern of the main effects is not restricted to entry-wise sparsity. Second, the data fitting term is not quadratic, but a heterogeneous exponential family quasi log-likelihood. This new framework enables us to tackle many more data sets combining  heterogeneous data, main effects and interactions. 

For the algorithmic development, our proposed method is related to the prior work such as \citep{lin2011linearized,tao2011recovering,Chen2015FastLE,pmlr-v51-gu16,DBLP:conf/aistats/ZhangWG18,jaggi2013revisiting,doi:10.1137/15M101628X, garber2018fast,lacoste2013block,beck2015cyclic,gidel2018frank}. These are based on various first-order optimization methods and shall be reviewed in detail in Section~\ref{algo}.  Among others, the MCGD method is mostly related to the recent FW-T method by \citet{doi:10.1137/15M101628X} that uses a mixed update rule to tackle a similar estimation problem. There are two differences: first, FW-T is focused on a quadratic loss which is a special case of the statistical estimation problem that we analyze; second, the per-iteration complexity of MCGD is lower as the update rules are simpler. Despite the simplifications, using a new proof technique, we prove that the convergence rate of MCGD is strictly faster than FW-T.

\paragraph{Notations:}
For any $m \in \NN$, $[m] \eqdef \{1,...,m\}$.
The operator ${\cal P}_\Omega (\cdot ):\mathbb{R}^{n\times p}\rightarrow \mathbb{R}^{n\times p}$ is the projection operator on the set of entries in $\Omega \subset [n]\times[p]$, and $(\cdot)_+ : \RR \rightarrow \RR_+$
is the projection operator on the non-negative orthant $(x)_+ \eqdef \max\{0,x\}$.
For matrices, we denote by $\norm{\cdot}_F$ the Frobenius norm, $\norm{\cdot}_\star$ the nuclear norm, $\norm{\cdot}$ the operator norm, and $\norm{\cdot}_{\infty}$ the entry-wise infinity norm.
For vectors, we denote by $\norm{\cdot}_1 $ is the $\ell_1$-norm, $\norm{\cdot}_2$ the Euclidean norm, $\norm{\cdot}_{\infty}$ the infinity norm, and $\norm{\cdot}_0$ the number of non zero coefficients.
The binary operator $\pscal{{\bm X}}{\bm Y}$ denotes the Frobenius inner product.
A function $f : \RR^q \rightarrow \RR$ is said to be $\sigma$-smooth if $f$ is continuously differentiable and 
$\| \grd f ( \prm ) - \grd f ( \prm' ) \|_2 \leq \sigma \| \prm - \prm' \|_2$ for all $\prm, \prm' \in \RR^q$.

%% file: problem.tex
\section{Problem Formulation}
\label{problem}
\paragraph{Heterogenous Data Model}  Let $(\Yset,\Xset)$ be a probability space equipped with a $\sigma$-finite measure $\mu$.  The canonical exponential family distribution $\{ \operatorname{Exp}_{h,g}(m), m \in \Xset\}$ with
base measure $h: \Yset \to \rset^+$, link function $g: \Xset \to \rset$,
and scalar parameter, $m \in \Xset$,  has  a density  given by
\beq
\label{exp-fam}
f_{m} (y) = h(y)\exp\left(y m  - g(m)\right) \,.
\eeq
The exponential family is a flexible framework to model different types of data.
For example, ${( \Yset = \RR, g(m)=m^2 \sigma^2/2, h(y)= (2 \pi \sigma^2)^{-1/2} \exp(-y^2/2\sigma^2) )}$
yields a Gaussian distribution with mean $m$ and variance $\sigma^2$ for numerical data;
${( \Yset = \{0,1\}, g(m)= \log(1+\exp(m)), h(y)= 1 )}$ yields a Bernoulli distribution with success probability $1/(1+\exp(-m))$ for binary data;
${( \Yset = \NN, g(m)= \exp(a m), h(y)= 1/y!)}$ where $a \in \rset$ yields a Poisson distribution with intensity $\exp(a m)$ for count data. In these cases, the parameter
space is $\Xset= \RR$.

Let $\{ (\Yset_j , g_j, h_j),~j \in [p] \}$ be a collection of observation spaces, base and link functions corresponding to the column types of a data frame $\mathbf{Y}= [\mathbf{Y}_{ij}]_{(i,j) \in [n] \times [p]} \in\Yset_1^{n}\times\ldots\times \Yset_p^{n}$. For each $i \in  [n]$ and $j \in  [p]$, we denote by $\mathbf{M}^0_{ij}$ the target parameter minimizing the Kullback-Leibler divergence between the distribution of $\mathbf{Y}_{ij}$ and the exponential family $\operatorname{Exp}_{h_j,g_j}$, $j \in [p]$, 
given by
\begin{equation}
\label{exp-model}
\mathbf{M}^0_{ij} = \argmax_m ~\mathbb{E}_{\mathbf{Y}_{ij}}[ \log(h_j(\mathbf{Y}_{ij})) + \mathbf{Y}_{ij}m - g_j(m)] \,.
\end{equation}
We propose the following model to estimate $\mathbf{M}^0 = [\mathbf{M}^0_{ij}]_{(i,j) \in [n] \times [p]}$ in the presence of additive effects and interactions.

\paragraph{LOw-rank Interaction with Sparse additive effects (LORIS) model}
For every entry $\mathbf{Y}_{ij}$, assume a vector of covariates $\mathbf{x}_{ij}\in\mathbb{R}^q$ is also available, e.g., user information and item characteristics. Denote $\mathbf{x}_{ij}(k)$, $k\in [q]$ the $k$-th component of $\mathbf{x}_{ij}$ and define the matrix $\mathbf{X}(k) = [\mathbf{x}_{ij}(k)]_{(i,j) \in [n] \times [p]}$.  We introduce the following decomposition of the parameter matrix $\mathbf{M}^0$:
\begin{equation}
\label{low-rank-model}
\mathbf{M}^0 = \sum_{k=1}^q\alp_k^0\mathbf{X}(k) + \Prm^0.
\end{equation}
We call \eqref{low-rank-model} the LORIS model, where $\alp\in \mathbb{R}^q$ is a sparse vector with unknown support modeling additive effects and $\mathbf{\Theta}^0\in\mathbb{R}^{n\times p}$ a low-rank matrix modeling the interactions.\\
In fact, LORIS is a generalization of \textit{robust} matrix completion \citep{Candes:2011:RPC}, where the parameter matrix can be decomposed as the sum of two matrices, one is low-rank and the other has some complementary low-dimensional structure such as entry-wise or column-wise sparsity.
Statistical recoverability results
in robust matrix estimation under a noiseless setting can be found in \citep{Xu:2010:RPV,Candes:2011:RPC, chandra, Hsu_robustmatrix}; the additive noise setting can be found in a recent work \citep{Klopp2017}.

\paragraph{Estimation Problem}
Denote $\Omega = \{(i,j)\in[n]\times [p] : \mathbf{Y}_{ij} \text{ is observed}\}$
as the observation set. For $\mathbf{M}\in\mathbb{R}^{n\times p}$, $\mathcal{L}(\mathbf{M})$ is the negative log-likelihood of the observed data $(\mathbf{Y}, \Omega)$ parameterized by $\mathbf{M}$. Up to an  additive constant,
\begin{equation}
\label{eq:log-likelihood}
\mathcal{L}(\mathbf{M}) = \sum_{(i,j)\in \Omega}\left\lbrace-\mathbf{Y}_{ij}\mathbf{M}_{ij}+g_j(\mathbf{M}_{ij})\right\rbrace \,.
\end{equation}
For $a > 0$, we consider the following estimation problem:
\begin{equation}
\label{eq:opt}
(\hat{\alp}, \hat{\Prm})\in\underset{\substack{\norm{\alp}_{\infty}\leq a\\\substack{\norm{\mathbf{\Theta}}_{\infty}\leq a}}}{\operatorname{argmin}} \quad \mathcal{L} \left(
\sum_{k=1}^q \alp_k {\bf X}(k) + \Prm
\right) + \lambda_S\norm{\alp}_1+\lambda_L\norm{\mathbf{\Theta}}_\star.
\end{equation}
We denote by $\hat{\bf M} = \sum_{k=1}^q \hat{\alp}_k {\bf X}(k) + \hat{\Prm}$
the estimated parameter matrix.
The $\ell_1$ and nuclear norm penalties are convex relaxations of the sparsity and low-rank constraints, and the regularization parameters $\lambda_S$ and $\lambda_L$ serve as trade-offs between fitting the data and enforcing sparsity of $\alp$ and controlling the "effective rank" of $\Prm$.
\paragraph{Statistical Guarantees}
\label{problem}
Here we establish  convergence rates for the joint estimation of $\alp^0$ and $\Prm^0$; the proofs can be found in the supplementary material. Consider the following assumptions.
\begin{assumption}
\label{ass:true-set}
$\norm{\Prm^0}_{\infty}\leq a$, $\norm{\alp^0}_{\infty}\leq a$ and for all $k\in[q]$ such that $\alp_k^0 \neq 0$, $\pscal{\Prm^0}{\mathbf{X}(k)} = 0$.
\end{assumption}
In particular, \Cref{ass:true-set} guarantees the uniqueness of the decomposition  in the LORIS model \eqref{low-rank-model}.
\begin{assumption}
\label{ass:dict} For $\nu> 0$, all $k\in [q]$ and $(i,j)\in [n]\times[p]$, $\mathbf{X}(k)_{ij}\in[-1,1].$ Furthermore for all $(i,j)\in [n]\times[p]$, $\sum_{k=1}^q|\mathbf{X}(k)_{ij}| \leq \nu.$
\end{assumption}
In particular, \Cref{ass:dict} guarantees that for all $(\Prm,\alp)$ satisfying  \Cref{ass:true-set}, the matrix $\mathbf{M} = \sum_{k=1}^q \alp_k {\bf X}(k) +\Prm$ satisfies $\norm{\mathbf{M}}_{\infty}\leq (1+\nu)a$.
Let $\mathbf{G}$ be the $q\times q$ Gram matrix of the dictionary $(\mathbf{X}(1),\ldots,\mathbf{X}(q))$ defined by $\mathbf{G} = [ \pscal{\mathbf{X}(k)}{\mathbf{X}(l)} ]_{ (k,l) \in [q] \times [q] }$.
\begin{assumption}
\label{ass:gram} For $\kappa>0$ and all $\alp \in \mathbb{R}^q$, $\alp^{\top} {\bf G} \alp \geq \kappa^2\norm{\alp}_2^2.$
\end{assumption}
Note we do not consider the case where the Gram matrix is singular, e.g., $q > np$. For $0<\sigma_-\leq \sigma_+< +\infty$ and $0<\gamma<\infty$ consider the following assumption on the link functions $g_j$:
\begin{assumption}
\label{ass:cvx}
The functions $g_j$ are twice differentiable, and for all $x\in [-(1+\nu)a-\gamma,(1+\nu)a+\gamma]$, 
$$\sigma_-^2\leq g_j''(x)\leq \sigma_+^2,~j \in [p].$$
\end{assumption}
\Cref{ass:cvx} implies the data fitting term $\mathcal{L}(\mathbf{M})$ is smooth and satisfies a restricted strong convexity property.
\begin{assumption}
\label{ass:subexp}
For all $(i,j)\in[n]\times[p]$, $Y_{ij}$ is a sub-exponential random variable with scale and variance parameters $1/\gamma$ and $\sigma_+^2$.
\end{assumption}
If the random variables $Y_{ij}$ are actually distributed according to an exponential family distribution of the form \eqref{exp-fam}, then \Cref{ass:cvx} implies \Cref{ass:subexp}.
\begin{assumption}
\label{ass:sampling-min}
For $(i,j) \in [n]\times [p]$, the events $\omega_{ij} = \{(i,j)\in\Omega\}$ are independent with occurrence probability   $\pi_{ij}$. Furthermore, there exists $0<\pi\leq 1$ such that for all $(i,j) \in [n]\times [p]$, $\pi_{ij}\geq \pi$.
\end{assumption}
\Cref{ass:sampling-min} implies a data missing-at-random scenario
where ${\bf Y}_{ij}$ is observed with probability at least $\pi$.
\begin{Theorem} \label{th:upper-bound}
Assume {H}\ref{ass:true-set}-\ref{ass:sampling-min}. Set
\beq
\lambda_L = 2C\sigma_+\sqrt{\pi\max(n,p)\log(n+p)}\text{,~~and}~~\lambda_S =  24\max_k\norm{\mathbf{X}(k)}_1\log(n+p)/\gamma,
\eeq
where $C$ is a positive constant. Assume that $\max(n,p)\geq 4\sigma_+^2/\gamma^6 \log^2(\sqrt{\min(n,p)/(\pi\gamma\sigma_-)}) + 2\exp(\sigma_+^2/\gamma^2+2\sigma_+^2\gamma a)$. Then, with probability at least $1 - 9(n+p)^{-1}$,
\begin{equation}
\label{cvg-rates}
\begin{aligned}
&\norm{\hat{\alp} - \alp^0}_2^2 && \leq C_1\frac{s\max_k\norm{\mathbf{X}(k)}_1\log(n+p)}{\kappa^2\pi} + \mathsf{D}_{\alp},\\
& \norm{\hat{\Prm} - \Prm^0}_F^2&&\leq C_2\left(\frac{r\max(n,p)}{\pi} + \frac{s\max_k\norm{\mathbf{X}(k)}_1}{\pi}\right)\log(n+p) + \mathsf{D}_{\Prm}.
\end{aligned}
\end{equation}
In \eqref{cvg-rates}, $s \eqdef \| \alp^0 \|_0$, $r \eqdef {\rm rank} (\Prm^0)$. $C_1$ and $C_2$ are positive constants and $\mathsf{D}_{\alp}$ and $\mathsf{D}_{\Prm}$ are residuals of lower order whose exact values are given in Appendix~A.
\end{Theorem}
The proof can be found in Appendix~A.
In \Cref{th:upper-bound}, the rate obtained for $\alp^0$ is the same as the bound obtained in \citep{Klopp2017} in the special case of robust matrix completion.
Examples satisfying $\max_k\norm{\mathbf{X}(k)}_1 / \kappa^2 = {\cal O}(1)$
include the case
where the elements of the dictionary are matrices are all zeros except a row or a column of one,
(to model row and column effects) and the number of rows $n$ and columns $p$ are of the same order;
or when the covariates ${\bf x}_{ij}$ are categorical and the categories are balanced, \ie the number of samples per category is of the same order.

The rate obtained for $\Prm^0$ is the sum of the standard low-rank matrix
completion rate of order $r\max(n,p)/\pi$, e.g., \citep{Klopp2014}, and
of a term which boils down to sparse vector estimation rate as long as
$\max_k\norm{\mathbf{X}(k)}_1 = {\cal O}(1)$.
Again, the latter can be satisfied by the special case of robust matrix completion, for which our rates match the results of \citep{Klopp2017}.

%% file: algorithm.tex

\section{A Mixed Coordinate Gradient Descent Method for LORIS}
\label{algo}
This section introduces a mixed coordinate gradient
descent (MCGD) method to solve the LORIS estimation problem \eqref{eq:opt}. 
We assume that $a$ is sufficiently large
such that the constraints $\| \alp \|_\infty \leq a, \| \Prm \|_\infty \leq a$ 
are always inactive.
To simplify notation, we denote the log-likelihood function
as ${\cal L}(\alp, \Prm) \eqdef \mathcal{L} \left(
\sum_{k=1}^q \alp_k {\bf X}(k) + \Prm
\right)$. 
We assume
\begin{assumption} \label{prop:L}
(a) ${\cal L}( \alp, \Prm )$ is  $\sigma_{\Prm}$-smooth \wrt $\Theta_{ij}$ for $(i,j) \in \Omega$ and (b) $\sigma_{\alp}$-smooth \wrt $\alp$; (c) the gradient $\grd_{\alp} {\cal L} ( \alp, \Prm )$ is $\hat{\sigma}_{\Prm}$-Lipschitz \wrt $\Prm$. Moreover, the gradient $\grd_{\Prm} {\cal L} ( \alp, \Prm )$ is bounded as long as $\alp, \Prm$
are bounded.
\end{assumption}
The above is implied by \Cref{ass:cvx} for 
bounded $(\alp, \Prm)$. 
We consider the augmented objective function:
\beq
F( \alp, \Prm, R ) \eqdef
{\cal L} ( \alp, \Prm ) + \lambda_S \| \alp \|_1 +  \lambda_L R \eqs.
\eeq
For some $R_{\sf UB} \geq 0$, if
an optimal solution $( \hat{\alp}, \hat{\Prm})$ to \eqref{eq:opt} satisfies
$\| \hat{\Prm} \|_\star \leq R_{\sf UB}$, then  any optimal solution to
the following problem
\beq \label{eq:opt_r}
{\sf P}(R_{\sf UB}):~~~~\min_{\alp \in \RR^q , \Prm \in \RR^{n \times p}, R \in \RR_+}~F( \alp, \Prm, R ) ~~{\rm s.t.}~~R_{\sf UB} \geq R \geq \| \Prm \|_\star \eqs,
\eeq
will also be optimal to \eqref{eq:opt}. For example, 
$( \hat{\alp}, \hat{\Prm}, \hat{R} )$
with $\hat{R} = \| \hat{\Prm} \|_\star$ is an optimal solution to \eqref{eq:opt_r}.   
We have defined the problem as ${\sf P}(R_{\sf UB})$ to
emphasize its dependence on the upper bound $R_{\sf UB}$.
Later we shall describe a simple strategy to estimate $R_{\sf UB}$.
We
fix the set $\Xi \subseteq [n] \times [p]$ where $\Omega \subseteq \Xi$
is the target coordinate set for the low rank matrix $\hat{\Prm}$ that we are interested
in.


\paragraph{Proposed Method}
A natural way to exploit structure in ${\sf P}( R_{\sf UB} )$ is to apply
coordinate gradient descent to update $\alp$ and $(\Prm,R)$ separately.
While the trace-norm constraint on $(\Prm,R)$ can be
handled by the conditional gradient (CG) method \citep{jaggi2013revisiting},
the $\ell_1$ norm penalization on $\alp$
is more efficiently tackled by the proximal gradient
method in practice.
In addition, we tighten the upper bound $R_{\sf UB}$ on-the-fly as
the algorithm proceeds.
The MCGD method goes as follows. At the $t$th iteration, we are given
the previous iterate $( \alp^{(t-1)} , \Prm^{(t-1)}, R^{(t-1)})$ and the upper bound $R_{\sf UB}^{(t)}$
is computed.
The first block $\alp$ is updated with a 
proximal gradient step:
  \beq \label{eq:prox}
  \begin{split}
  \alp^{(t)} & = {\rm prox}_{ \gamma \lambda_S \| \cdot \|_1 } \big( \alp^{(t-1)} - \gamma \grd_{\alp} {\cal L} ( \alp^{(t-1)}, \Prm^{(t-1)}) \big) \\
  & = {\sf T}_{\gamma \lambda_S} \big( \alp^{(t-1)} - \gamma \grd_{\alp} {\cal L} ( \alp^{(t-1)}, \Prm^{(t-1)})  \big) \eqs.
  \end{split}
  \eeq
In \eqref{eq:prox}, $\grd_{\alp} {\cal L} (\cdot)$ is the gradient of the log-likelihood function
taken \wrt 
$\alp$, 
$\gamma > 0$ is a pre-defined step size parameter and
${\sf T}_\lambda ( {\bm x} ) \eqdef {\rm sign} ( {\bm x} ) \odot ( {\bm x} - \lambda {\bf 1} )_+$
is the component-wise soft thresholding operator.
Alternatively, we can exactly solve the problem
\beq \label{eq:exact_pg} \textstyle
\alp^{(t)} \in \argmin_{ \alp \in \RR^q }~F( \alp, \Prm^{(t-1)}, R^{(t-1)} ) \eqs,
\eeq
for which closed-form solution can be obtained in certain special cases (see below).

The second block $(\Prm,R)$ is updated with a
CG step
\beq \label{eq:fw}
(\Prm^{(t)} , R^{(t)}) = ( \Prm^{(t-1)}, R^{(t-1)} ) + \beta_t ( \hat{\Prm}^{(t)} - \Prm^{(t-1)},\hat{R}^{(t)} - R^{(t-1)}) \eqs,
\eeq
where $\beta_t \in [0,1]$ is a step size to be defined later. 
$(\hat{\Prm}^{(t)}, \hat{R}^{(t)})$ is a direction evaluated as
\beq \label{eq:fw_dir}
(\hat{\Prm}^{(t)},\hat{R}^{(t)}) \in \argmin_{ {\bm Z}, R } ~\langle {\bm Z}, \grd_{\Prm} {\cal L}( \alp^{(t)}, \Prm^{(t-1)}  ) \rangle + \lambda_1 R ~~{\rm s.t.}~~ \| {\bm Z} \|_\star \leq R \leq R_{\sf UB}^{(t)} \eqs,
\eeq
and $\grd_{\Prm} {\cal L}(\cdot)$ is the gradient of ${\cal L}(\cdot)$ 
taken \wrt $\Prm$.
If $(\Prm^{(t-1)}, R^{(t-1)})$ is feasible to ${\sf P}( R_{\sf UB}^{(t)} )$,
then $( \Prm^{(t)}, R^{(t)} )$ must also be feasible to ${\sf P}( R_{\sf UB}^{(t)} )$.
Furthermore, if we let ${\bm u}_1, {\bm v}_1$ be the top left and right
singular vectors of the gradient matrix
$\grd_{\Prm} {\cal L}( \alp^{(t)}, \Prm^{(t-1)}  )$ and
$\sigma_1(\grd_{\Prm} {\cal L}( \alp^{(t)}, \Prm^{(t-1)}  ))$  be the top singular value,
then $(\hat{\Prm}^{(t)}, \hat{R}^{(t)})$ admits a simple closed form solution:
\beq \label{eq:dt}
(\hat{\Prm}^{(t)}, \hat{R}^{(t)}) = \begin{cases}
({\bm 0}, 0), & \text{if}~\lambda_L \geq \sigma_1(\grd_{\Prm} {\cal L}( \alp^{(t)}, \Prm^{(t-1)}  ) ) \eqs,  \\
(-R_{\sf UB}^{(t)} {\bm u}_1 {\bm v}_1^\top, R_{\sf UB}^{(t)}), & \text{if}~\lambda_L < \sigma_1( \grd_{\Prm} {\cal L}( \alp^{(t)}, \Prm^{(t-1)}  ) ) \eqs.
\end{cases}
\eeq
Lastly, the step size $\beta_t$ is determined by:
  \beq \label{eq:beta}
  \beta_t = \min \Big\{ 1, \frac{ \langle \Prm^{(t-1)} - \hat{\Prm}^{(t)} , \grd_{\Prm} {\cal L} ( \alp^{(t)}, \Prm^{(t-1)}  )  \rangle + \lambda_L (R^{(t-1)} - \hat{R}^{(t)}  ) }{ \sigma_{\Prm} \| {\cal P}_\Omega ( \hat{\Prm}^{(t)} - \Prm^{(t-1)} ) \|_{\rm F}^2 }  \Big\} \eqs.
  \eeq
The step size strategy ensures
 decrease in the objective value between successive iterations.
This is essential for establishing
convergence of the proposed method [cf.~Theorem~\ref{prop:main}].
We remark that the arithmetics in the MCGD method are not affected
when we restrict the update of $\Prm^{(t)}$ 
in \eqref{eq:fw} to the entries in $\Xi$ only.
This is due to ${\cal L}( {\bm X})
= {\cal L} ( {\cal P}_\Omega ( {\bm X} ) )$ and the CG update
direction \eqref{eq:fw_dir}
only involves the gradient of $\grd_{\Prm} {\cal L}( \alp^{(t)}, \Prm^{(t-1)}  ) $
\wrt entries of $\Prm$ in $\Omega$, where $\Omega \subseteq \Xi$.
\paragraph{Computing the Upper Bound $R_{\sf UB}^{(t)}$} We describe a strategy
for computing a valid upper bound $R_{\sf UB}^{(t)}$
for $\hat{R}$ and $\| \hat{\Prm} \|_\star$ during the updates in the MCGD method.
Let us assume that:
\begin{assumption} \label{ass:lb}
For all $\Prm$ and $\alp$, we have ${\cal L} ( \alp, \Prm ) \geq 0$.
\end{assumption}
The above can be enforced as   the log-likelihood function is lower bounded [cf.~\Cref{ass:cvx}].
From \eqref{eq:opt} and using the above assumption, it is obvious that
\beq
F_0( {\bm 0}, {\bm 0} ) = {\cal L} ( {\bm 0} , {\bm 0} ) \geq   {\cal L} ( \hat{\alp}, \hat{\Prm}) + \lambda_S \| \hat{\alp}\|_1 + \lambda_L \| \hat{\Prm} \|_\star \geq \lambda_L \| \hat{\Prm} \|_\star,
\eeq
and thus $R_{\sf UB}^0 \eqdef \lambda_L^{-1}  {\cal L} ( {\bm 0} + f_{\bm U}({\bm 0}))$
is a valid upper bound to $\| \hat{\Prm} \|_\star$; furthermore it can be tightened
as we progress in the MCGD method.
In particular, observe that $(\hat{\alp}, \hat{\Prm}, \hat{R})$ with
$\hat{R} = \| \hat{\Prm} \|_\star$ is an
optimal solution to ${\sf P}( R_{\sf UB}^0 )$,
we have
\beq
F( \alp, \Prm, R ) \geq F(\hat{\alp}, \hat{\Prm}, \hat{R}) = {\cal L} ( \hat{\alp}, \hat{\Prm}  ) + \lambda_S \| \hat{\alp}\|_1 + \lambda_L \hat{R} \geq \lambda_L \hat{R}.
\eeq
In other words, for all feasible $( \alp, \Prm, R )$ to ${\sf P}( R_{\sf UB}^0)$,
$\lambda_L^{-1} F( \alp, \Prm, R )$ is  an upper bound
to $\hat{R}$ and $\| \hat{\Prm} \|_\star$.
The above motivates us to select
$R_{\sf UB}^{(t)} \eqdef  \lambda_L^{-1} F( \alp^{(t)}, \Prm^{(t-1)} , R^{(t-1)} )$
at iteration $t$, where we   observe that
$R_{\sf UB}^{(t)} \geq R^{(t-1)}$. That is, $(\alp^{(t)}, \Prm^{(t-1)}, R^{(t-1)})$ is feasible 
to both ${\sf P}( R_{\sf UB}^{(t)} )$ and ${\sf P}( R_{\sf UB}^{(t-1)} )$.
Lastly, we summarize the MCGD method in Algorithm~\ref{alg:lrm_fw}.
\vspace{-.1cm}
\begin{multicols}{2}{
\textbf{Computation Complexity}~~Consider the   MCGD method  in Algorithm~\ref{alg:lrm_fw}.
Observe that line~\ref{line:inexact} requires computing the gradient \wrt $\alp$ which
involves $|\Omega| q$ Floating Points Operations (FLOPS)
and the soft thresholding operator involves ${\cal O}(q)$ FLOPS.
As the log-likelihood function ${\cal L} (\cdot)$ is evaluated element-wisely on $\Prm$,
evaluating the objective value and
the derivative \wrt $\Prm$ requires ${\cal O}(| \Omega |)$ FLOPS.
As such,
line~\ref{line:eval} can be evaluated in ${\cal O}(|\Omega|)$ FLOPS
and line~\ref{line:fw} requires ${\cal O}( | \Omega | \max\{ n, p \} \log (1 / \delta) )$ FLOPS
where the additional complexity is due to the top SVD computation
and $\delta$ is a preset accuracy level of SVD computation.
Lastly, 
line~\ref{line:end} requires ${\cal O}( | \Xi | )$ FLOPS
since we only need to update the entries of $\Prm$ in $\Xi$ [cf.~see the remark
after \eqref{eq:beta}].
The overall per-iteration complexity   is
${\cal O} ( |\Xi| + |\Omega| ( \max\{ n, p \} \log ( 1/ \delta) + q ) )$.
\algsetup{indent=0.75em}
\begin{algorithm}[H]
\caption{MCGD Method for \eqref{eq:opt_r}.}\label{alg:lrm_fw}
  \begin{algorithmic}[1]
  \STATE \textbf{Initialize:} --- $\Prm^{(0)}, \alp^{(0)}, R^{(0)}$. E.g., $\Prm^{(0)}, \alp^{(0)}, R^{(0)} = ( {\bm 0}, {\bm 0}, 0 )$.
  \FOR {$t=1,2,\dots,T$}
  \STATE \label{line:inexact} \texttt{\emph{// Update for $\alp$ //}}\\
   Compute the proximal update using \eqref{eq:prox} [or exact update via \eqref{eq:exact_pg}] to obtain
  $\alp^{(t)}$.
  \STATE \label{line:eval} \texttt{\emph{// Update for $(\Prm,R)$ //}}\\
  Compute the upper bound as $R_{\sf UB}^{(t)} \eqdef  \lambda_L^{-1} F( \alp^{(t)}, \Prm^{(t-1)} , R^{(t-1)} )$.
  \STATE \label{line:fw} Compute the update direction, $(\hat{\Prm}^{(t)}, \hat{R}^{(t)})$, using Eq.~\eqref{eq:dt}.
  \STATE Compute the CG update using \eqref{eq:fw}, where the step size $\beta_t$ is set as Eq.~\eqref{eq:beta}. \label{line:end}
\ENDFOR
\STATE \textbf{Return:} $\Prm^{(T)}, \alp^{(T)}, R^{(T)}$.
  \end{algorithmic}
\end{algorithm}} \end{multicols}\vspace{-.2cm}
 From the above, the per-iteration computation complexity
of the MCGD method scales linearly with the problem dimension
$\max\{ n, p \}$ and $ |\Omega|$. This is comparable to \citep{doi:10.1137/15M101628X,garber2018fast}, where the former focuses
only on the least square loss case.
The following theorem, whose proof can be found in Appendix C, shows that the MCGD method
converges at a sublinear rate.
\begin{Theorem} \label{prop:main}
Assume H\ref{prop:L} and H\ref{ass:lb}. 
Define the  quantity
\beq \label{eq:Ct}
C(t) \eqdef \max \Big\{ \frac{24 (Q^{(t)})^2}{\gamma}, \frac{24 \hat{\sigma}_{\Prm}^2 (Q^{(t)})^2}{\sigma_{\Prm}} + \max\{ 6 R_{\sf UB}^{(t)} ( \lambda_L + M^{(t)}), 24 \sigma_{\Prm} (R_{\sf UB}^{(t)})^2 \} \Big\} \eqs,
\eeq
where $Q^{(t)} \eqdef \lambda_S^{-1} F( \alp^{(t)}, \Prm^{(t)}, R^{(t)} )$,
$M^{(t)} \eqdef \| \grd_{\Prm} {\cal L} ( \alp^{(t)}, \Prm^{(t-1)} ) \|_2$
and $R_{\sf UB}^{(t)} \eqdef \lambda_L^{-1} F( \alp^{(t)}, \Prm^{(t-1)}, R^{(t-1)})$.
If we choose the step sizes as $\gamma \leq 1 / \sigma_{\alp}$ and $\beta_t$ as in \eqref{eq:beta}, then (i) the above quantity is upper bounded
as $C(t) \leq \overline{C}$ for all
$t \geq 1$, where
\beq
\overline{C} \eqdef \max \Big\{ \frac{24 (Q^{(0)})^2}{\gamma}, \frac{24 \hat{\sigma}_{\Prm}^2 (Q^{(0)})^2}{\sigma_{\Prm}} + \max\{ 6 R_{\sf UB}^{(0)} ( \lambda_L + \bar{M}), 24 \sigma_{\Prm} (R_{\sf UB}^{(0)})^2 \} \Big\} \eqs,
\eeq
such that $\bar{M}$ is an upper bound to $M^{(t)}$,
and (ii)
the MCGD method converges to an $\epsilon$-optimal
solution to \eqref{eq:opt}
in $T$ iterations, \ie
$F_0( \alp^{(T)}, \Prm^{(T)} ) - F_0 ( \hat{\alp}, \hat{\Prm} ) \leq \epsilon$, where
\beq \label{eq:requiredno}
T  \geq \overline{C}(T) \Big( \frac{1}{\epsilon} - \frac{1}{F_0( \alp^{(0)}, \Prm^{(0)} ) - F_0(\hat{\alp}, \hat{\Prm})} \Big)_+ ~~\text{with}~~\overline{C}(T) \eqdef \Big( \frac{1}{T} \sum_{t=1}^T \frac{1}{C(t)} \Big)^{-1} \eqs.
\eeq
In particular, as $\overline{C}(T) \leq  \overline{C}$, at most
$\overline{C} ( \epsilon^{-1} - (F_0( \alp^{(0)}, \Prm^{(0)} ) - F_0(\hat{\alp}, \hat{\Prm}))^{-1} )_+$ iterations are required
for the MCGD method to reach an $\epsilon$-optimal solution
to \eqref{eq:opt}.
\end{Theorem}
\paragraph{Detailed Comparison to Prior Algorithms} 
Previous contributions have focused on the special case of \eqref{eq:opt} 
where $q = np$, the dictionary $(\mathbf{X}(1),\ldots,\mathbf{X}(q))$ is the canonical basis of $\mathbb{R}^{n\times p}$, and the link functions are quadratic. In this particular case, \eqref{eq:opt} becomes the estimation problem solved in sparse plus low-rank matrix decomposition.
Popular examples are the alternating direction method of multiplier
\citep{lin2011linearized,tao2011recovering}
or the projected gradient method on a reformulated problem
\citep{Chen2015FastLE}. These methods either require computing a complete 
SVD    or knowing the optimal rank number of $\Prm$ a priori.  
When $n,p \gg 1$, it is computationally prohibitive to evaluate the complete SVD since each iteration would require ${\cal O}( \max\{ n^2 p , p^2 n \})$ FLOPS.
Other related work rely on factorizing the low-rank component, yielding
nonconvex problems \citep{pmlr-v51-gu16}; see also
\citep{DBLP:conf/aistats/ZhangWG18} and references therein. 

Similar to the development of MCGD, a natural alternative is to apply 
algorithms based on
the CG (a.k.a.~Frank-Wolfe) method \citep{jaggi2013revisiting},
whose iterations only require the computation of a top SVD. 
The present work is closely related to the efforts in \citep{doi:10.1137/15M101628X, garber2018fast} which focused on the quadratic setting. \citet{doi:10.1137/15M101628X} combines the CG method with proximal update 
as a two-steps procedure; \citet{garber2018fast} combines a CD method 
with CG updates on both the sparse and low-rank components.
The work in \citep{garber2018fast} is also related to 
\citep{lacoste2013block,beck2015cyclic} which combine CD with CG updates
for solving constrained problems, instead of penalized problems like \eqref{eq:opt}. 
Sublinear convergence rates are proven for the above methods. 
Finally, \citet{fithian2013scalable} also suggested to apply CD on \eqref{eq:opt},
yet the convergence properties were not discussed.

In fact, when the MCGD's result is specialized to the same setting as
\citep{doi:10.1137/15M101628X}, our worst-case bound on iteration number
computed with $\overline{C}$ match 
the bound in \citep{doi:10.1137/15M101628X}.
As shown in the supplementary material, we have $C(t) \rightarrow C^\star$,
where $C^\star$ depends on the optimal objective value of \eqref{eq:opt_r}
and is smaller than $\overline{C}$.
Since the quantity $\overline{C}(T)$ in \eqref{eq:requiredno}
is an average of  $\{ C(t) \}_{t = 1}^T$, 
this implies that the MCGD method requires less number of iterations for convergence
than that is required by \citep{doi:10.1137/15M101628X}.
Such reduction is possible due to the on-the-fly update for $R_{\sf UB}^{(t)}$.
Moreover, our analysis in Theorem~\ref{prop:main} holds when the MCGD method is implemented with a few practical modifications. 


\paragraph{Exact Partial Minimization for $\alp$}
Consider the special case of \eqref{eq:opt} where the link functions are either 
quadratic or exponential and the
dictionary matrices satisfy:
\beq
{\rm supp} ( {\bm X}(k) ) \cap {\rm supp} ( {\bm X}(k') ) = \emptyset,~k \neq k'~~\text{and}~~
[ {\bm X}(k) ]_{i,j} = c_k,~\forall~(i,j) \in {\rm supp} ( {\bm X}(k) ) \eqs.
\eeq
In this case, the 
partial minimization 
\eqref{eq:exact_pg} 
can be decoupled into $q$ scalar optimizations involving 
one coordinate of $\alp$, which can be solved in closed form.
Note that this modification
to the MCGD method is supported by
Theorem~\ref{prop:main} and the sublinear convergence rate holds. 
On the contrary, closed form update of $\alp$ is not supported by
prior works such as \citep{doi:10.1137/15M101628X, garber2018fast, 
lacoste2013block,beck2015cyclic}.

\paragraph{Distributed MCGD Optimization}
Consider the case where the observed data entries are stored across $K$ workers, each of them communicating with a central server.
It is natural to distribute the MCGD optimization over these workers
to offload computation burden, or for privacy protection. 
Formally, we divide $\Omega$ into $K$ disjoint partitions
such that $\Omega = \Omega_1 \cup 
\cdots \cup \Omega_K$ and worker $k$ holds $\Omega_k$.
In this way,
${\cal L} ( \alp, \Prm ) = \sum_{k=1}^K {\cal L}_k (\alp, \Prm)$, where
${\cal L}_k ( \alp, \Prm )$ is defined by replacing the summation over $\Omega$
with $\Omega_k$ in \eqref{eq:log-likelihood}. 
Clearly, when $\alp$ and ${\cal P}_{\Omega_k} (\Prm)$ are given to the $k$th worker, the 
worker will be able to evaluate the \emph{local} loss function and its gradient. \\
As shown in Appendix D, 
the MCGD method can be easily extended to utilize distributed computation. 
The proximal update in line~\ref{line:inexact} is replaced by the following procedure. First,
the \emph{local} gradients computed by the workers are aggregated, then the soft thresholding
operation is performed at the central server.
Meanwhile, as the CG update in line~\ref{line:fw} essentially requires computing 
the top singular vectors of the gradient matrix 
$\grd_{\Prm} L(\alp,\Prm) = \sum_{k=1}^K \grd_{\Prm} {\cal L}_k(\alp,{\cal P}_{\Omega_k} (\Prm)) $,
the latter can be implemented through a distributed
version of the power method exploiting the decomposable structure 
of the gradient, such as described in \citep{zheng2017distributed}. 
It only requires ${\cal O}( \log(1/\delta))$ 
power iterations to compute a top SVD solution of accuracy $\delta$. 
Thus, for a sufficiently small $\delta>0$, 
the overall per-iteration complexity of the distributed
method at the $t$th iteration
is reduced to ${\cal O}( | \Xi | + \max\{n,p\} \log(1/\delta) )$ at the central server, and ${\cal O}( |\Omega_k| ( \max\{n,p\} \log (1/\delta) +  q) )$ at the $k$th worker.

%% file: sim.tex

\section{Numerical Experiments}
\label{simus}
\paragraph{Experimental Setup} 
We first generate the target parameter ${\bf M}^0$ according to the 
LORIS model in \eqref{low-rank-model}. 
For the sparse additive effects component, we consider $q = pn / 5$ 
where 
we set $( {\bf X}(k) )_{ij} = 1$ if $j(n-1) + i \in \{ 5(k-1) + 1,..., 5k \}$.
This models a 
categorical variable containing $n/5$ categories. 
Furthermore, the target sparse component $\alp^0$ has a sparsity level of $10\%$.
For the low-rank component, the target parameter $\Prm^0$ 
is generated as a rank-$4$ matrix formed by the outer product 
of random orthogonal vectors.
Notice that due to the structure of sparse additive effects, the surveyed prior methods \citep{lin2011linearized,pmlr-v51-gu16,Chen2015FastLE} cannot be applied directly.

\paragraph{Gaussian Design}
To compare our framework to a reasonable benchmark, we focus on a homogenous  setting with numerical data modeled with the quadratic link function $g(m) = m^2$. We set the regularization parameters $\lambda_S$ and $\lambda_L$ to the theoretical values given in \Cref{th:upper-bound}. We compare our result with a common two-step procedure where the components $\alp_{kj}$ are first estimated in a preprocessing step as the means of the variables taken by group; then $\Prm$ is estimated using the softImpute method proposed in \citep{softImpute}. The regularization parameter for \citep{softImpute} is set to the same value $\lambda_L$. We compare the results in terms of estimation error and computing time in \Cref{simu-gaus}, after letting the two methods converge to the same precision of $10^{-5}$. We observe the two methods perform equally well in terms of estimating $\Prm$. LORIS yields constant estimation errors of $\alp^0$ as the dimension increases and the support of $\alp^0$ is kept constant, contrary to the two-step procedure for which the estimation error of $\alp^0$ increases with the dimension. 
As expected, the two-step method is faster for small data sets, whereas for large data sizes LORIS is superior in computational time. The above results are consistent with our theoretical findings.\vspace{-.1cm}
\begin{table}[ht]
\centering
\begin{tabular}{l l l l l l l l l}
  \toprule
\textbf{problem size ($n \times p$)} & \multicolumn{2}{c }{\textbf{time} (secs)}
& \multicolumn{2}{c}{$\norm{\Prm^0-\hat\Prm}_F^2$} & \multicolumn{2}{c}{$\norm{\alp^0-\hat\alp}_2^2$}\\
  \midrule
   & LORIS & two-step & LORIS & two-step &LORIS & two-step  \\
  \midrule
$150\times 30$ & $0.17$ & $0.02$  & $52$ & $52$ &$1.8$ & $3.0$ \\
  \hline
$1,500\times 300$ & $13.8$ & $10.7$ &  $175.5$ & $234$ & $0.95$ & $17.1$ \\
\hline
$15,000\times 300$ & $130.2$ & $136.6$ & $675$ & $720$ & $0.95$ & $16.2$ \\
\hline
$15,000\times 3,000$ & $348$ & $528$ & $2.7\times 10^{3}$ & $2.6\times 10^{3}$ & $2.34$ & $180$ \\
\bottomrule
\end{tabular}\vspace{.1cm}
\caption{Comparison of proposed method with a two-step method in terms of computation time and estimation error for increasing dimensions (averaged over $10$ experiments).}\vspace{-.5cm}
\label{simu-gaus}
\end{table} 
\paragraph{Survey data}
To test the efficacy of our framework with heterogeneous data, 
we examine a survey conducted by the French National Institute of Statistics (Insee: \texttt{http://www.insee.fr/}) concerning the hobbies of
French people. The data set contains $n=8,403$ individuals and $p=19$ binary and quantitative variables, indicating 
whether or not the person has been involved in different activities (reading, fishing, etc.), the number of hours spent watching TV and the overall number of hobbies of the individuals. Individuals are grouped by age category ($15-25$, $25-35$, etc.): this categorical variable is used as a predictor of the survey responses in the subsequent experiment.
\begin{figure}[ht]
\begin{subfigure}[b]{0.4\textwidth}
\hspace{-1cm}
\includegraphics[scale = 0.4]{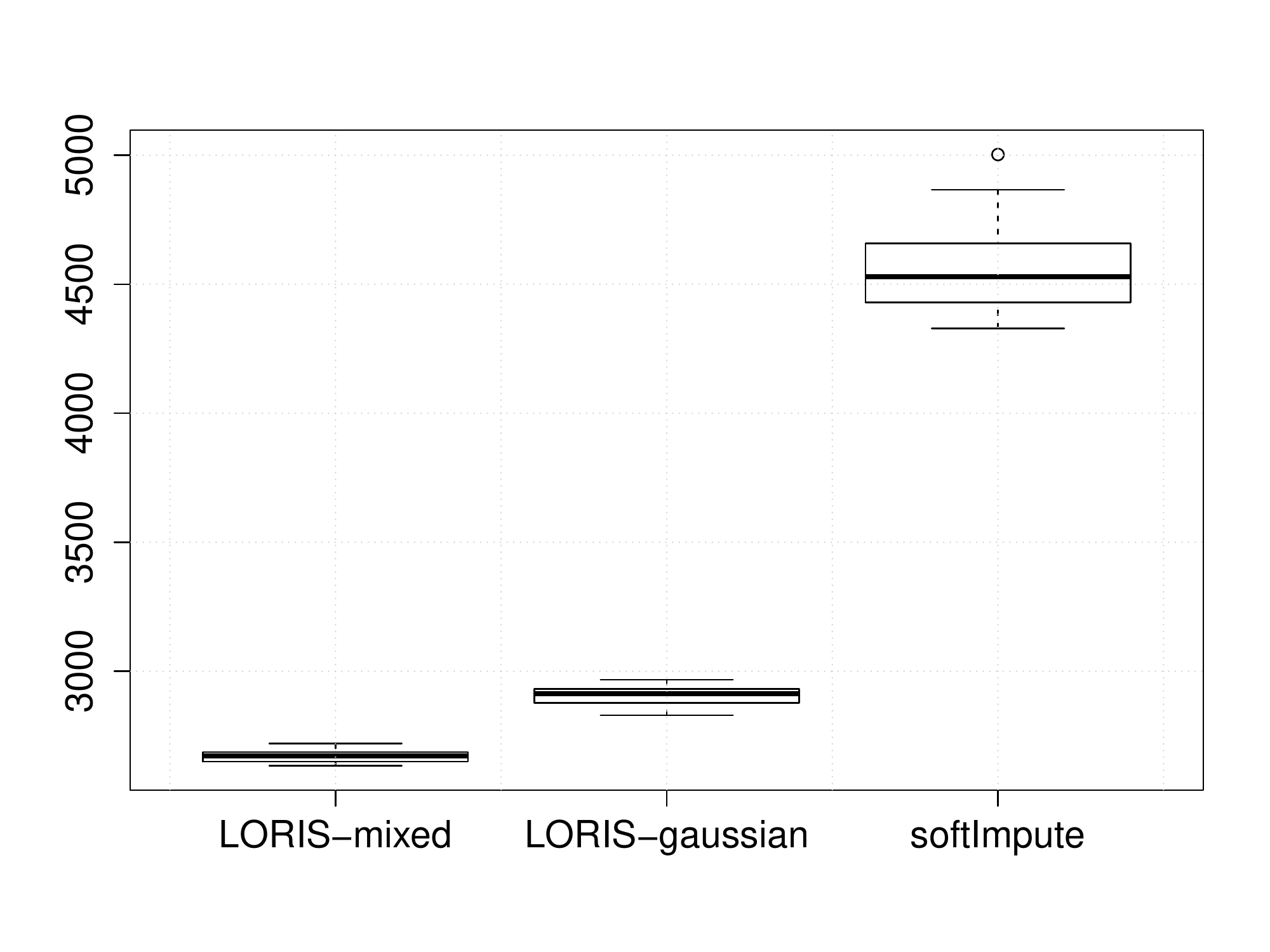}
\end{subfigure}
\hspace{1cm}
\begin{subfigure}[b]{0.4\textwidth}
\includegraphics[scale = 0.4]{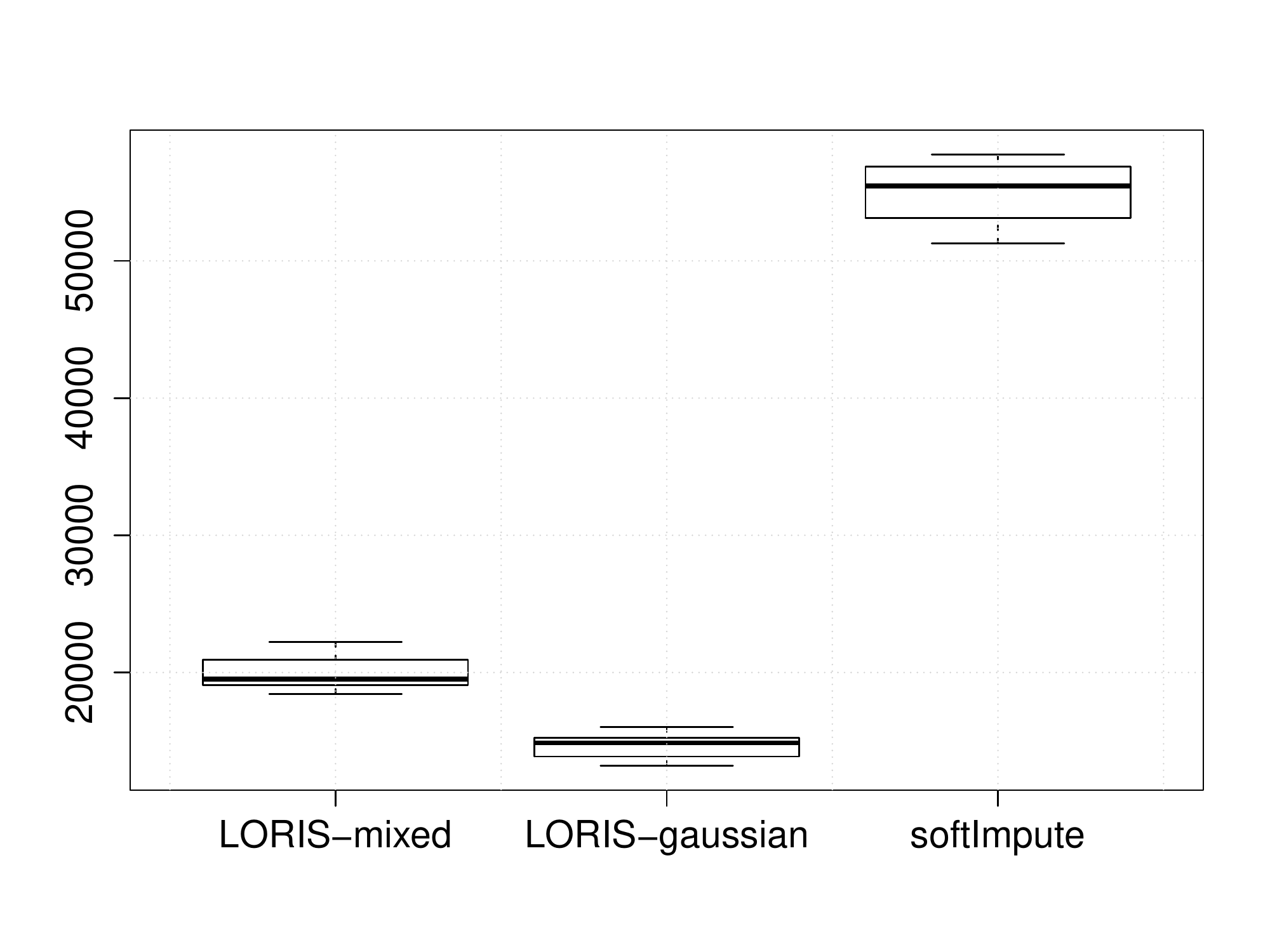}
\end{subfigure}
\captionof{figure}{Imputation error of LORIS with mixed data model and Gaussian data model, and softImpute (10 replications) for categorical variables (left) and quantitative variables (right).}\vspace{-.2cm}
\label{imputation}
\end{figure}
We introduce $30$\% of missing values in the data set, and compare the imputation error of LORIS with a mixed data model (using a quadratic loss for numeric columns, a logistic loss for binary columns and a Poisson loss for counts) and LORIS with a Gaussian data model, with the imputation error of softImpute. The results are given in \Cref{imputation} across $10$ replications of the experiment, and show that, for this example, both LORIS models improve on the baseline softImpute  by a factor $2$. We also observe that modeling explicitly the binary variables leads to better imputation. 

Finally, we apply LORIS with a mixed data model to the original data set. A subset of the resulting $\alp$ vector is given in \Cref{alpha}. There is a coefficient in $\alp_{kj}$ for every age category $k$ and every variable $j$.  
The coefficients in \Cref{alpha} indicate that young individuals engage in activities such as music and sport more than older people, and the opposite trend for collecting, knitting and fishing. Some coefficients are set to zero, indicating the absence of effect of the age category on the variable.
We also observe that younger people engage overall in more activities than older people.
\begin{table}[ht]
\centering
\begin{tabular}{llllllll}
  \toprule
Age category & Music & Sport & Collecting & Mechanic & Knitting & Fishing & Nb activities \\ 
  \midrule
  25-35 & 2.2 & 0.4 & -2.1 & 0 & -1.7 & -1.9 & 10.0 \\ 
  35-45 & 2.0 & 0.3 & -2.7 & 0 & -2.3 & -2.3 & 13.0 \\ 
  45-55 & 1.1 & -0.8 & -2.1 & 0 & -2.7 & -2.7 & 13.8 \\ 
  55-65 & 0 & -2.2 & -1.9 & 0 & -1.0 & -1.6 & 8.8 \\ 
  65-75 & 0 & -2.1 & -1.4 & -1.1 & -0.7 & -1.3 & 5.5 \\ 
  75-85 & -0.1 & -0.9 & -0.6 & -0.5 & -0.1 & -0.6 & 2.2 \\ 
   \bottomrule
\end{tabular}\vspace{.1cm}
\caption{Estimated age category effects ($\alp$).}\vspace{-.5cm}
\label{alpha}
\end{table}

\paragraph{Conclusion} In this paper, we proposed a new framework
for handling large data frames with heterogeneous data and missing values which incorporates additive effects. It consists of a doubly penalized quasi-maximum likelihood estimator
and a new optimization algorithm to implement the estimator. 
We examined both the statistical and computational efficiency of 
the framework and derived worst case bounds of its performance. 
Future work includes the incorporation of qualitative features with more than two categories and of missing values in the dictionary matrices.

%% file: append_stats_guarantees.tex
\appendix

\section{Statistical guarantees}
\subsection{Main result}
We recall the convergence rates for the Frobenius norm of the errors $\Delta \Prm = \hat \Prm - \Prm^0$ and $\Delta \alp = \hat\alp - \alp^0$ given in \Cref{problem}. 
Define $d_{\mathbf{X}} = \max_k\norm{\mathbf{X}(k)}_1$ and the following quantities:
\begin{equation*}
\label{eq:upper-bounds-rates2-alpha}
\mathsf{D}_{\alp} = \frac{\norm{\alp^0}_1}{\pi}\frac{\log (n+p)}{\sigma_-^2\gamma} +\left(\frac{a}{\pi}\right)^2\log (n+p),
\end{equation*}
\begin{equation*}
\label{eq:upper-bounds-rates2-L}
\mathsf{D}_{\Prm} = \mathsf{D}_{\alp} +  d_{\mathbf{X}}\norm{\alp^0}_1\left\lbrace\frac{12\pi\sqrt{\log(n+p)}}{\gamma(1+\nu)a\sigma_+ \sqrt{\beta}}+\frac{1}{\pi\sigma_-^2}\left(\frac{\log (n+p)}{\gamma}\right)+1\right\rbrace .
\end{equation*}
We assume that $M = (n\vee p)$ is large enough, that is
$$M\geq  \left\lbrace\frac{4\sigma_+^2}{\gamma^6}\log^2\left(\frac{\sqrt{n\wedge p}}{p\gamma\sigma_-}\right)\vee 2\exp\left(\sigma_+^2/\gamma^2\vee\sigma_+^2\gamma(1+\nu a)\right)\right\rbrace.$$
\begin{Theorem} \label{th:upper-bound-app}
Assume {H}\ref{ass:true-set}-\ref{ass:sampling-min}. Set
\beq
\lambda_L = 2C\sigma_+\sqrt{\pi\max(n,p)\log(n+p)}\text{,~~and}~~\lambda_S\geq 24\max_k\norm{\mathbf{X}(k)}_1\log(n+p)/\gamma,
\eeq
where $C$ is a positive constant. Assume that $\max(n,p)\geq 4\sigma_+^2/\gamma^6 \log^2(\sqrt{\min(n,p)/(\pi\gamma\sigma_-)}) + 2\exp(\sigma_+^2/\gamma^2+2\sigma_+^2\gamma a)$. Then, with probability at least $1 - 9(n+p)^{-1}$,
\begin{equation}
\label{cvg-rates}
\begin{aligned}
&\norm{\hat{\alp} - \alp^0}_2^2 && \leq C_1\frac{sd_{\mathbf{X}}\log(n+p)}{\kappa^2\pi} + \mathsf{D}_{\alp},\\
& \norm{\hat{\Prm} - \Prm^0}_F^2&&\leq C_2\left(\frac{r\max(n,p)}{\pi} + \frac{sd_{\mathbf{X}}}{\pi}\right)\log(n+p) + \mathsf{D}_{\Prm}.
\end{aligned}
\end{equation}
In \eqref{cvg-rates}, $s \eqdef \| \alp^0 \|_0$, $r \eqdef {\rm rank} (\Prm^0)$. $C_1$ and $C_2$ are positive constants and $\mathsf{D}_{\alp}$ and $\mathsf{D}_{\Prm}$ are residuals of lower order whose exact values are given in Appendix~A.
\end{Theorem}
Denoting by $\lesssim$ the inequality up to constant and logarithmic factors, the order of magnitude of the bounds are therefore:
\begin{equation*}
\begin{aligned}
& \norm{\Delta \alp}_2^2 && \lesssim \frac{sd_{\mathbf{X}}}{p\kappa^2},\\
& \norm{\Delta \Prm}_F^2 && \lesssim  \frac{r\beta}{p^2}+ \frac{sd_{\mathbf{X}}}{p},
\end{aligned}
\end{equation*}
where $s = \norm{\alp^0}_0$ and $r = \operatorname{rank}(\Prm^0)$. In the case of almost uniform sampling, \textit{i.e.} $c_1 \pi \leq \pi_{ij}\leq c_2 \pi$ for all $(i,j)\in[n]\times[p]$ and two positive constants $c_1$ and $c_2$, we obtain that $\beta \leq c_2 (n\vee p)\pi$, which yields the following simplified bound: 
\begin{equation}
\label{simplified-bound}
\norm{\Delta \Prm}_F^2\lesssim \frac{rM}{\pi} + \frac{sd_{\mathbf{X}}}{\pi}.
\end{equation}
The rate given in \eqref{simplified-bound} is the sum of the usual low-rank convergence rate $rM/p$ and, when $d_{\mathbf{X}}$ is a constant, of the usual sparse vector convergence rate.

\subsection{Sketch of the proof}

Let $\{\epsilon_{ij}\}$ be an i.i.d. Rademacher sequence independent of $Y$ and $\Omega$. We define
$$\Sigma_R = \sum_{i=1}^{n}\sum_{j=1}^{p}\omega_{ij}\epsilon_{ij}E_{ij}.$$ 
In \Cref{th:general-th} we give a general result under some assumptions on the regularization parameters $\lambda_L$ and $\lambda_S$, which depend on the random matrices $\nabla\mathcal{L}(\mathbf{M}^0)$ and $\Sigma_R$. Then, \Cref{Lemma:SigmaR} and \ref{Lemma:Sigma} allow us to compute values of $\lambda_L$ and $\lambda_S$ that satisfy the assumptions of \Cref{th:general-th} with high probability. Finally we combining these results yield \Cref{th:upper-bound-app}. Define
\begin{equation}
\label{eq:upper-bounds-rate-alpha}
\Psi_{\alp} = \frac{\norm{\alp^0}_1}{\pi}\left\lbrace \frac{\lambda_S}{\sigma_-^2} + a^2d_{\mathbf{X}}\EE{\norm{\Sigma_R}_{\infty}}\right\rbrace+\left(\frac{a}{\pi}\right)^2\log(n+p),
\end{equation}
\begin{equation}
\label{eq:upper-bounds-rate-L}
\Psi_{\Prm}  = \frac{r}{\pi^2}\EE{\norm{\Sigma_R}}^2 + \frac{\norm{\alp}_1}{\pi}\left\lbrace\frac{\lambda_S}{(1+\nu)a\lambda_L}+d_{\mathbf{X}}\EE{\norm{\Sigma_R}_{\infty}} \right \rbrace
+ \Psi_{\alp}.
\end{equation}

\begin{Theorem}
\label{th:general-th}
Let $$\lambda_L\geq 2\norm{\nabla\mathcal{L}(\mathbf{M}^0)}, \quad \lambda_S\geq 2d_{\mathbf{X}}\left(\norm{\nabla\mathcal{L}(\mathbf{M}^0)}_{\infty}+ 2\sigma_+^2(1+\nu)a\right),$$
and assumptions \textbf{H}~\ref{ass:dict}-\ref{ass:sampling-min} hold. Then, with probability at least $1-8(n+p)^{-1}$
\begin{equation}
\begin{aligned}
& (i)\quad &&\norm{\Delta \alp}_2^2\leq \frac{C}{\kappa^2}\Psi_{\alp}\text{, and}\\
& (ii)\quad && \norm{\Delta \Prm}_F^2 \leq  C\left\lbrace\frac{r\lambda_L^2}{\pi^2\sigma_-^4} +(1+\nu)a \Psi_{\Prm}\right\rbrace.
\end{aligned}
\end{equation}
\end{Theorem}

Denote $\Delta \mathbf{M} = \hat{\mathbf{M}} -\mathbf{M}^0$. We first derive an upper bound on the Frobenius error restricted to the observed entries $\norm{\mathcal{P}_{\Omega}(\Delta \mathbf{M})}_F^2$. Then we show some restricted strong convexity property, meaning that $\EE\norm{\mathcal{P}_{\Omega}(\Delta \mathbf{M})}_F^2$ is upper bounded by $\norm{\mathcal{P}_{\Omega}(\Delta \mathbf{M})}_F^2$ up to a residual term defined later.
\paragraph{Upper bound on $\norm{\mathcal{P}_{\Omega}(\Delta \mathbf{M})}_F^2$.}
By definition of $\hat{\Prm}$ and $\hat{\alp}$:
\begin{equation*}
\mathcal{L}(\hat{\mathbf{M}}) -  \mathcal{L}(\mathbf{M}^0)
\leq \lambda_L\left(\norm{\Prm^0}_\star-\norm{\hat{\Prm}}_\star \right) + \lambda_S\left(\norm{\alp^0}_1 - \norm{\hat{\alp}}_1 \right).
\end{equation*}
Recall that, for $\alp\in\mathbb{R}^q$, we use the notation $f_U({\alp}) = \sum_{k=1}^q\alp_k\mathbf{X}(k).$
Adding $\pscal{\nabla  \mathcal{L}(\mathbf{M}^0)}{\Delta \mathbf{M}}$ on both sides of the last inequality, we get
\begin{multline*}
\mathcal{L}(\hat{\mathbf{M}}) -  \mathcal{L}(\mathbf{M}^0)+\pscal{\nabla  \mathcal{L}(\mathbf{M}^0)}{\Delta \mathbf{M}} \leq 
\lambda_L\left(\norm{\Prm^0}_\star-\norm{\hat{\Prm}}_\star \right)-\pscal{\nabla  \mathcal{L}(\mathbf{M}^0)}{\Delta \Prm}\\
  +\lambda_S\left(\norm{\alp^0}_1 - \norm{\hat{\alp}}_1\right) - \pscal{\nabla  \mathcal{L}(\mathbf{M}^0)}{f_U(\Delta \alp)}.
\end{multline*}
The strong convexity of the link functions $g_j$, $j\in[p]$, allows us to lower bound the left hand side term and obtain
\begin{multline*}
\frac{\sigma_-^2}{2}\norm{\mathcal{P}_{\Omega}(\Delta \mathbf{M} )}_F^2 \leq \lambda_L\left(\norm{\Prm^0}_\star-\norm{\hat{\Prm}}_\star \right)-\pscal{\nabla  \mathcal{L}(\mathbf{M}^0)}{\Delta \Prm}\\
  +\lambda_S\left(\norm{\alp^0}_1 - \norm{\hat{\alp}}_1\right) - \pscal{\nabla  \mathcal{L}(\mathbf{M}^0)}{f_U(\Delta \alp)}.
\end{multline*}
We now upper bound the right hand side using the following three agruments: the duality of the norms $\norm{\cdot}_\star$ and $\norm{\cdot}$ on the one hand and of the norms $\norm{\cdot}_1$ and $\norm{\cdot}_{\infty}$ on the other hand, the triangular inequality and the following assumptions:
$$\lambda_L\geq 2\norm{\nabla  \mathcal{L}(\mathbf{M}^0)}, \quad \lambda_S \geq 2\norm{\nabla\mathcal{L}(\mathbf{M}^0)}_{\infty}d_{\mathbf{X}}.$$
We obtain
\begin{equation}
\label{eq:ineq_x_omega}
\norm{\mathcal{P}_{\Omega}(\Delta \mathbf{M})}_F^2\leq \frac{3\lambda_L}{\sigma_-^2}\sqrt{2\operatorname{rank}(\mathbf{M}^0)}\norm{\Delta \Prm}_F+ \frac{3\lambda_S}{\sigma_-^2}\norm{\alp^0}_1.
\end{equation}
\paragraph{Restricted strong convexity}
We now show that when the errors $\Delta \Prm$ and $\Delta \alp$ belong to a subspace $\mathcal{C}$ and for a residual $\mathsf{D}$ - both defined later on - the following holds with high probability:
\begin{equation}
\label{eq:res-str-cvx}
\norm{\mathcal{P}_{\Omega}(\Delta \mathbf{M})}_F^2\geq \EE\norm{\mathcal{P}_{\Omega}(\Delta \mathbf{M})}_F^2 -\mathsf{D}.
\end{equation}
We start by defining the set $\mathcal{C}$ and prove that it contains the errors $\Delta \Prm$ and $\Delta \alp$ with high probability (Lemma \ref{Lemma:alpha-l1}-\ref{Lemma:L-nuc}); then we show that restricted strong convexity holds on this subspace (\Cref{Lemma:rsc}).\\

For non-negative constants $d_1$, $d_{\Pi}$, $\rho < m$ and $\varepsilon$ that will be specified later on, define the two following sets:
\begin{equation}
\label{eq:sets-A}
\mathcal{A}(d_1,d_{\Pi}) =  \left\lbrace \alp\in\mathbb{R}^{q}\text{ : } \norm{\alp}_1\leq d_1\text{, }\norm{\mathcal{P}_{\Omega}(f_U(\alp))}_F^2\leq d_{\Pi}\right\rbrace.
\end{equation}
The constants $d_1$ and $d_{\Pi}$ define the constraints on the $\ell_1$ norm of $\alp$ and weighted Frobenius norm of $f_U(\alp)$.
\begin{equation}
\label{eq:sets-C}
\begin{aligned}
& \mathcal{L}(\rho,\varepsilon) && = \Bigg\lbrace \Prm\in\mathbb{R}^{n\times p}, \alp\in\RR^q: \norm{\mathcal{P}_{\Omega}(\Prm+f_U(\alp))}_F^2\geq \frac{72\log(n+p)}{\pi\log(6/5)},\\
& && \quad \quad \quad \quad\quad \quad \quad \quad
\norm{\Prm+f_U(\alp)}_{\infty}\leq 1,\norm{\Prm}_\star\leq \sqrt{\rho}\norm{\Prm}_F + \varepsilon \Bigg\rbrace
\end{aligned}
\end{equation}
Condition $\norm{\Prm}_\star\leq \sqrt{\rho}\norm{\Prm}_F + \varepsilon$ is a relaxed form of the condition $\norm{\Prm}_\star\leq \sqrt{\rho}\norm{\Prm}_F$ satisfied for matrices of rank $\rho$.
Finally, we define the constrained set of interest:
$$\mathcal{C}(d_1,d_{\Pi},\rho,\varepsilon) = \mathcal{L}(\rho,\varepsilon) \cap \left\lbrace \mathbb{R}^{n\times p}\times \mathcal{A}(d_1,d_{\Pi}) \right\rbrace.$$
Let
\begin{equation*}
\begin{aligned}
& d_1 && = 4\norm{\alp}_1,\\
& d_{\Pi} && = \frac{3\lambda_S}{\sigma_-^2}\norm{\alp^0}_1 + 64a^2d_{\mathbf{X}}\EE{\norm{\Sigma_R}_{\infty}}\norm{\alp}_1 + 3072a^2\pi^{-1} + \frac{72a^2\log(n+p)}{\log(6/5)}.
\end{aligned}
\end{equation*}
The following Lemma, proved in \Cref{Lemma:alpha-l1-proof} states that with high probability, $\Delta\alp\in\mathcal{A}(d_1,d_{\Pi})$.
\begin{Lemma}
\label{Lemma:alpha-l1}
Let $\lambda_S\geq 2d_{\mathbf{X}}\left(\norm{\nabla\mathcal{L}(\mathbf{M}^0)}_{\infty}+ 2\sigma_+^2(1+d_{\mathbf{X}})a\right)$ and assume \textbf{H}~\ref{ass:dict}-\ref{ass:sampling-min} hold. Then, with probability at least $1-8(n+p)^{-1}$,
$$\Delta\alp \in \mathcal{A}(d_1,d_{\Pi});$$
\end{Lemma}
\Cref{Lemma:alpha-l1} (proved in \Cref{Lemma:L-nuc-proof}) implies $(i)$ of \Cref{th:general-th}. Thus, we only need to prove $(ii)$.
\begin{Lemma}
\label{Lemma:L-nuc}
Let 
$$\lambda_L\geq 2\norm{\nabla\mathcal{L}(\mathbf{M}^0)},\quad \lambda_S\geq 2d_{\mathbf{X}}\left(\norm{\nabla\mathcal{L}(\mathbf{M}^0)}_{\infty}+ 2\sigma_+^2(1+d_{\mathbf{X}})a\right),$$
and assumption \textbf{H}~\ref{ass:cvx} hold. Then, for $\rho = 32r$ and $\varepsilon = 3\lambda_S/\lambda_L\norm{\alp^0}_1$,
$$
\norm{\Delta \Prm}_\star \leq \sqrt{\rho}\norm{\Delta \Prm}_F + \varepsilon.$$
\end{Lemma}
A proof of \Cref{Lemma:L-nuc} can be found in \Cref{Lemma:L-nuc-proof}. As a consequence, under the conditions on the regularization parameters $\lambda_L$ and $\lambda_S$ given in \Cref{Lemma:L-nuc} and whenever $$\EE\norm{\mathcal{P}_{\Omega}(\Delta \Prm+f_U({\Delta\alp}))}_F^2\geq \frac{72\log(n+p)}{\pi\log(6/5)},$$ the error terms $(\Delta \Prm,\Delta \alp)$ belong to the constrained set $\mathcal{C}(d_1,d_{\Pi},\rho,\varepsilon)$ with high probability. We therefore consider the two possible cases: $\EE\norm{\mathcal{P}_{\Omega}(\Delta \Prm+f_U({\Delta\alp}))}_F^2<\frac{72\log(n+p)}{\pi\log(6/5)}$ and $\EE\norm{\mathcal{P}_{\Omega}(\Delta \Prm+f_U({\Delta\alp}))}_F^2\geq\frac{72\log(n+p)}{\pi\log(6/5)}$.

\textbf{Case 1:} Suppose $\EE\norm{\mathcal{P}_{\Omega}(\Delta \Prm+f_U({\Delta\alp}))}_F^2<\frac{72\log(n+p)}{\pi\log(6/5)}$. Then, \Cref{Lemma:alpha-l1} combined with the fact that $\norm{\mathbf{M}}_F^2\leq \pi^{-1}\norm{\mathcal{P}_{\Omega}(\mathbf{M})}_F^2$ for all $\mathbf{M}$, and the identity $(a+b)^2\geq a^2/4-4b^2$ ensures that
$$\norm{\Delta \Prm}_F^2 \leq 4\norm{\Delta \Prm + f_U(\Delta \alp)}_F^2 + 16 \norm{f_U(\Delta\alp)}_F^2,$$
therefore
$$\norm{\Delta \Prm}_F^2 \leq \frac{288a^2\log(n+p)}{\log(6/5)} + 16 \Phi_{\alp},$$
which implies (ii) of \Cref{th:general-th}.\\

\textbf{Case 2:}
 Suppose $\EE\norm{\mathcal{P}_{\Omega}(\Delta \Prm + f_U(\Delta \alp))}_F^2\geq\frac{72\log(n+p)}{\pi\log(6/5)}$. Then, \Cref{Lemma:alpha-l1}  and \ref{Lemma:L-nuc} yield that with probability at least $1-8(n+p)^{-1}$,
$$\left(\frac{\Delta \Prm}{2(1+\nu)a},\frac{\Delta \alp}{2(1+\nu)a}\right)\in\mathcal{C}(d_1',d_{\Pi}',\rho', \varepsilon'), \text{ with}$$
\begin{equation*}
\begin{aligned}
& d_1' && = \frac{d_1}{2(1+\nu)a},\quad
& d_{\Pi}' && = \frac{d_{\Pi}}{4(1+\nu)^2a^2},\\
& \rho' && = \rho, & \varepsilon' && = \frac{\varepsilon}{2(1+\nu)a},
\end{aligned}
\end{equation*}
where $d_1,d_{\Pi},\rho$ and $\varepsilon$ are defined in \Cref{Lemma:alpha-l1}  and \ref{Lemma:L-nuc}. We use the following result, proved in \Cref{Lemma:rsc-proof}. Define the set $\tilde{\mathcal{A}}(d_1)$ as follows:
\begin{equation*}
\label{eq:set-A}
\tilde{\mathcal{A}}(d_1) = \left\lbrace \alp\in\mathbb{R}^q:\quad \norm{\alp}_{\infty}\leq 1;\quad \norm{\alp}_1\leq d_1;\quad \norm{\mathcal{P}_{\Omega}(f_U{\alp})}_F^2\geq \frac{18\log(n+p)}{\pi\log(6/5)}\right\rbrace.
\end{equation*}
Let $d_1$, $d_{\Pi}$, $\rho$ and $\varepsilon$ be positive constants, and 
\begin{equation}
\label{eq:residuals}
\begin{aligned}
& \mathsf{D}_{\alp} && = 8 \nu d_1 d_{\mathbf{X}}\EE{\norm{\Sigma_R}_{\infty}}+768\pi^{-1},\\
& \mathsf{D}_{X} && = \frac{112\rho}{\pi}\EE{\norm{\Sigma_R}}^2 + 8\nu \varepsilon \EE{\norm{\Sigma_R}}+8\nu d_1d_{\mathbf{X}}\EE{\norm{\Sigma_R}_{\infty}} + d_{\Pi}+768\pi^{-1}.
\end{aligned}
\end{equation}
\begin{Lemma}
\label{Lemma:rsc}
Assume \textbf{H}~\ref{ass:sampling-min}. Then, the following properties hold:
\begin{enumerate}[label=(\roman*)]
\item For any $\alp\in\tilde{\mathcal{A}}(d_1)$, with probability at least $1-8(n+p)^{-1}$,
$$\norm{\mathcal{P}_{\Omega}(f_U(\alp))}_F^2\geq \frac{1}{2}\EE\norm{\mathcal{P}_{\Omega}(f_U(\alp))}_F^2- \mathsf{D}_{\alp}.$$
\item For any pair $(\Prm,\alp)\in \mathcal{C}(d_1,d_{\Pi},\rho,\varepsilon)$, with probability at least $1-8(n+p)^{-1}$
\begin{equation}
\norm{\mathcal{P}_{\Omega}(\Delta \Prm+f_U({\Delta\alp}))}_F^2\geq \frac{1}{2}\EE\norm{\mathcal{P}_{\Omega}(\Delta \Prm+f_U({\Delta\alp}))}_F^2- \mathsf{D}_{X}.
\end{equation}
\end{enumerate}
\end{Lemma}
\Cref{Lemma:rsc} is proved in \Cref{Lemma:rsc-proof}. We apply \Cref{Lemma:rsc} (ii) to $\left(\frac{\Delta \Prm}{2(1+\nu)a},\frac{\Delta \alp}{2(1+\nu)a}\right)$ which implies that with probability at least $1-8(n+p)^{-1}$,
$\mathbb{EE}\norm{\mathcal{P}_{\Omega}(\Delta \mathbf{M})}_F^2 \leq 2\norm{\mathcal{P}_{\Omega}(\Delta \mathbf{M})}_F^2 + 2(1+\nu)a\Psi_{\Prm} $. Combined with \eqref{eq:ineq_x_omega} and $\norm{\Delta \mathbf{M}}_F^2\leq \pi^{-1}\EE\norm{\mathcal{P}_{\Omega}(\Delta \mathbf{M})}_F^2$, it implies that
\begin{equation*}
\norm{\Delta \mathbf{M}}_F^2\leq \frac{6\sqrt{2r}\lambda_L}{p\sigma_-^2}\norm{\Delta \Prm}_F +\frac{6\lambda_S}{\pi
\sigma_-^2}\norm{\alp^0}_1+2(1+\nu)a\Psi_{\Prm}.
\end{equation*}
Now using $\norm{\Delta \mathbf{M}}_F^2\geq \frac{\norm{\Delta \Prm}_F^2}{2} - \norm{f_U(\Delta\alp) }_F^2$ and
$\frac{6\sqrt{2r}\lambda_L}{\pi\sigma_-^2}\norm{\Delta \Prm}_F \leq \frac{\norm{\Delta \Prm}_F^2}{4} + \frac{288r\lambda_L^2}{p^2\sigma_-^4}$, we obtain
$$\norm{\Delta \Prm}_F^2\leq \frac{1152r\lambda_L^2}{p^2\sigma_-^4} + \frac{24\lambda_S\norm{\alp^0}_1}{\pi\sigma_-^2} +2(1+\nu)a \Psi_{\Prm}+4\Psi_{\alp},$$
which gives the result of \Cref{th:general-th} (ii).

We now give deterministic upper bounds on $\EE{\norm{\Sigma_R}}$ and $\EE{\norm{\Sigma_R}_{\infty}}$, and probabilistic upper bounds on $\norm{\nabla\mathcal{L}(\mathbf{M}^0)}$ and $\norm{\nabla\mathcal{L}(\mathbf{M}^0)}_{\infty}$. We will use them to select values of $\lambda_L$ and $\lambda_S$ which satisfy the assumptions of \Cref{th:general-th} and compute the corresponding upper bounds. 

\begin{Lemma}{\citep[Lemma 10]{Klopp2017}}
\label{Lemma:SigmaR}
Let assumption \textbf{H}~\ref{ass:sampling-min} hold. Then, there exists an absolute constant $C^\star$ such that the two following inequalities hold
\begin{equation*}
\label{eq:sigmaR-infty}
\EE{\norm{\Sigma_R}_{\infty}} \leq 1,\text{ and}
\end{equation*}
\begin{equation*}
\label{eq:sigmaR-nuc}
\EE{\norm{\Sigma_R}} \leq C^\star\left\{\sqrt{\beta} + \sqrt{\log (\min(n,p))}\right\}.
\end{equation*}
\end{Lemma}

\begin{Lemma}{\cite[Lemma 10]{Klopp2017}}
\label{Lemma:Sigma}
Let assumptions \textbf{H}~\ref{ass:true-set}-\ref{ass:sampling-min} hold. Then, there exists an absolute constant $c^\star$ such that the following two inequalities hold with probability at least $1-(n+p)^{-1}$.
\begin{equation}
\label{eq:Sigma-infty}
\norm{\nabla\mathcal{L}(\mathbf{M}^0)}_{\infty}\leq 6\max \left\{\sigma_+\sqrt{\log (n+p)}, \frac{\log (n+p)}{\gamma}\right\},
\end{equation}

\begin{equation}
\label{eq:Sigma-nuc}
\norm{\nabla\mathcal{L}(\mathbf{M}^0)}\leq c^\star\max\left\lbrace \sigma_+\sqrt{\beta\log (n+p)}, \frac{\log (n+p)}{\gamma }\log\left(\frac{1}{\sigma_-}\sqrt{\frac{np}{\beta}}\right) \right\rbrace.
\end{equation}
\end{Lemma}

From \Cref{th:general-th}, \Cref{Lemma:SigmaR} and \ref{Lemma:Sigma} combined with a union bound argument, we deduce result given in \Cref{problem}.

\section{Technical results}

\subsection{Proof of \Cref{Lemma:alpha-l1}}
\label{Lemma:alpha-l1-proof}
We start by proving $\norm{\Delta\alp}_1\leq 4\norm{\alp^0}_1$. By the optimality conditions over a convex set \cite[Chapter 4, Section 2, Proposition 4]{AubinEkeland}, there exist two subgradients $\hat{f}_{\Prm} $ in the subdifferential of $\norm{\cdot}_\star$ taken at $\hat{\Prm}$ and $\hat{f}_{\alp}$ in the subdifferential of $\norm{\cdot}_1$ taken at $\hat{\alp}$, such that for all feasible pairs $(\Prm,\alp)$ we have

\begin{equation}
\label{eq:opt-cond}
\pscal{\nabla\mathcal{L}(\hat{\mathbf{M}})}{\Prm-\hat{\Prm}+\sum_{k=1}^q(\alp_k-\hat{\alp}_k)\mathbf{X}(k)}+\lambda_L\pscal{\hat{f}_{\Prm}}{\Prm-\hat{\Prm}}+\lambda_S\pscal{\hat{f}_{\alp}}{\alp-\hat{\alp}}\geq 0.
\end{equation}
Applying inequality \eqref{eq:opt-cond} to the pair $(\hat{\Prm},\alp^0)$ we obtain
$$\pscal{\nabla\mathcal{L}({\hat{\mathbf{M}}})}{\sum_{k=1}^q\Delta\alp_k\mathbf{X}(k)}+\lambda_S\pscal{\hat{f}_{\alp}}{\Delta\alp}\geq 0.
 $$
Denote $\tilde {\mathbf{M}} = \hat{\Prm} + \sum_{k=1}^q \alp^0_k\mathbf{X}(k)$. The last inequality is equivalent to 
\begin{multline*}
\underbrace{\pscal{\diff{\TX}}{f_U(\Delta\alp)}}_{\mathsf{B}_1}+\underbrace{\pscal{ \diff{\tilde{\mathbf{M}}}-\diff{\TX}}{f_U(\Delta\alp)}}_{\mathsf{B}_2}+\underbrace{\pscal{\diff{\hat{\mathbf{M}}}- \diff{\tilde{\mathbf{M}}}}{f_U(\Delta\alp)}}_{\mathsf{B}_3}\\+\lambda_S\pscal{\hat{f}_{\alp}}{\Delta\alp}\geq 0.
\end{multline*}
We now derive upper bounds on the three terms $\mathsf{B}_1$, $\mathsf{B}_2$ and $\mathsf{B}_3$ separately. Recall that we denote $\umax = \max_k\norm{\mathbf{X}(k)}_{1}$ and bound $\mathsf{B}_1$ as follows:
\begin{equation}
\label{eq:boundI}
\mathsf{B}_1\leq \norm{\Delta\alp}_{1}\norm{\diff{\TX}}_{\infty}\umax.
\end{equation}
Similarly, the duality between $\linf$ and $\lone$ gives 
$$\mathsf{B}_2\leq \norm{\Delta \alp}_{1}\norm{ \diff{\tilde{\mathbf{M}}}-\diff{\TX}}_{\infty}\umax.$$
Moreover,  $\diff{\tilde{\mathbf{M}}}-\diff{\TX}$ is a matrix with entries $g_j'(\tilde{\mathbf{M}}_{ij})-g_j'(\TX_{ij})$, therefore assumption \textbf{H}~\ref{ass:cvx} ensures 
$$\norm{\diff{\tilde{\mathbf{M}}}-\diff{\TX}}_{\infty}\leq 2\smax^2(1+\nu)a, $$
and finally we obtain
\begin{equation}
\label{eq:boundII}
\mathsf{B}_2\leq \norm{\Delta \alp}_{1}2\smax^2(1+\nu)a\umax .
\end{equation}
We finally bound $\mathsf{B}_3$ as follows. We have that
$$\mathsf{B}_3 = \sum_{i=1}^{n}\sum_{j=1}^{p}\omega_{ij}\left(g_j'(\hat{\mathbf{M}}_{ij})-g_j'(\tilde{\mathbf{M}}_{ij})\right)\left(\tilde{\mathbf{M}}_{ij}-\hat{\mathbf{M}}_{ij}\right).$$
Now, for all $j\in\nint{1}{p}$, $g_j'$ is increasing therefore
$$\left(g_j'(\hat{\mathbf{M}}_{ij})-g_j'(\tilde{\mathbf{M}}_{ij})\right)\left(\tilde{\mathbf{M}}_{ij}-\hat{\mathbf{M}}_{ij}\right)\leq 0,$$ which implies $\mathsf{B}_3\leq 0.$
 Combined with \eqref{eq:boundI} and \eqref{eq:boundII} this yields
$$\lambda_S\pscal{\hat{f}_{\alp}}{\hat\alp - \Talpha}\leq  \norm{\Delta\alp}_{1}\umax\left(\norm{\diff{\TX}}_{\infty}+ 2\smax^2(1+\nu)a\right).$$
Besides, the convexity of $\lone$ gives $\pscal{\hat{f}_{\alp}}{\hat\alp - \Talpha}\geq \norm{\hat{\alp}}_{1}-\norm{\Talpha}_{1}$, therefore
\begin{multline*}
\left\lbrace\lambda_S-\umax\left(\norm{\diff{\TX}}_{\infty}+ 2\smax^2(1+\nu)a\right)\right\rbrace\norm{\hat\alp}_{1}\leq\\
\left\lbrace\lambda_S+\umax\left(\norm{\diff{\TX}}_{\infty}+ 2\smax^2(1+\nu)a\right)\right\rbrace\norm{\Talpha}_{1},
\end{multline*}
and the condition $\lambda_S\geq 2\left\lbrace\umax\left(\norm{\diff{\TX}}_{\infty}+ 2\smax^2(1+\nu)a\right)\right\rbrace$ gives $\norm{\hat \alp}_{1}\leq 3\norm{\Talpha}_{1}$ and finally
\begin{equation}
\label{eq:ineq-alpha-l1}
\norm{\Delta\alp}_{1}\leq 4\norm{\Talpha}_{1}.
\end{equation}

We consider the two following cases.
\paragraph{Case 1:} $\EE\norm{\mathcal{P}_{\Omega}(\fu{\Delta\alp})}_F^2 < \frac{72a^2\log(n+p)}{\pi\log(6/5)}$. Then the result holds trivially.
\paragraph{Case 2:} $\EE\norm{\mathcal{P}_{\Omega}(\fu{\Delta\alp})}_F^2 \geq \frac{72a^2\log(n+p)}{\pi\log(6/5)}$. For $d_1>0$ recall the definition of the set 
\begin{equation*}
\label{eq:set-A}
\tilde{\mathcal{A}}(d_1) = \left\lbrace \alp\in\R^q:\quad \norm{\alp}_{\infty}\leq 1;\quad \norm{\alp}_{1}\leq d_1;\quad \EE\norm{\mathcal{P}_{\Omega}(\fu{\Delta\alp})}_F^2\geq \frac{18\log(n+p)}{\pi\log(6/5)}\right\rbrace.
\end{equation*}
Inequality \eqref{eq:ineq-alpha-l1} and $\norm{\Delta\alp}_{\infty}\leq 2a$ imply that 
$$\frac{\Delta \alp}{2a} \in \tilde{\mathcal{A}}\left(\frac{2\norm{\Talpha}_{1}}{a} \right).$$
Therefore we can apply \Cref{Lemma:rsc}(i) and obtain that with probability at least $1-8(n+p)^{-1}$,
\begin{equation}
\label{eq:upper-bound-alpha-subres}
\EE\norm{\mathcal{P}_{\Omega}(\fu{\Delta\alp})}^2\leq 2\norm{\mathcal{P}_{\Omega}(\fu{\Delta\alp})}_F^2 + 64\nu a\norm{\Talpha}_{1}\umax\pe{\norm{\Sigma_R}_{\infty}} + 3072a^2p^{-1}.
\end{equation}
We now must upper bound the quantity $\norm{\mathcal{P}_{\Omega}(\fu{\Delta\alp})}_F^2$. Recall that $\tilde{\mathbf{M}} = \sum_{k=1}^q\Talpha_k\mathbf{X}(k)+\hat{\mathbf{M}}$. By definition, $$\mathcal{L}(\hat{\mathbf{X}})+ \lambda_L\norm{\hat{\Prm}}_{\star} + \lambda_S\norm{\hat{\alp}}_{1}\leq \mathcal{L}(\tilde{\mathbf{M}})+ \lambda_L\norm{\hat{\Prm}}_{\star} + \lambda_S\norm{\Talpha}_{1},$$ i.e.
\begin{equation*}
\mathcal{L}(\hat{\mathbf{M}})- \mathcal{L}(\tilde{\mathbf{M}})  \leq 
\lambda_S\left(\norm{\Talpha}_{1}-\norm{\hat{\alp}}_{1}\right).
\end{equation*}
Substracting $\pscal{\nabla\mathcal{L}(\tilde{\mathbf{M}})} {\hat{\mathbf{M}} - \tilde{\mathbf{M}}}$ on both sides and by strong convexity of $\mathcal{L}$ we obtain
\begin{equation}
\label{eq:th2-subres}
\begin{aligned}
& \frac{\smin^2}{2}\norm{\mathcal{P}_{\Omega}(\fu{\Delta \alp})}^2 && \leq \lambda_S\left(\norm{\Talpha}_{1}-\norm{\hat{\alp}}_{1}\right) +\pscal{\nabla  \mathcal{L}(\tilde{\mathbf{M}})}{\fu{\Delta\alp}}\\
& && \leq \lambda_S\left(\norm{\Talpha}_{1}-\norm{\hat{\alp}}_{1}\right) + \underbrace{\left|\pscal{\nabla  \mathcal{L}(\TX)}{\fu{\Delta \alp}}\right|}_{\mathsf{C}_1}\\
& && +\underbrace{\left|\pscal{\nabla  \mathcal{L}(\TX)-\nabla  \mathcal{L}(\tilde{\mathbf{M}})}{\fu{\Delta \alp}}\right|}_{\mathsf{C}_2}
\end{aligned}.
\end{equation}
The duality of $\norm{\cdot}_{1}$ and $\norm{\cdot}_{\infty}$ yields $\mathsf{C}_1 \leq  \norm{\nabla  \mathcal{L}(\TX)}_{\infty}d_{\mathbf{X}}\norm{\Delta\alp}_{1}$, and
$$\mathsf{C}_2 \leq  \norm{\nabla  \mathcal{L}(\TX)-\nabla  \mathcal{L}(\tilde{\mathbf{M}})}_{\infty}\umax\norm{\Delta \alp}_{1}.$$
Furthermore, $$\norm{\nabla  \mathcal{L}(\TX)-\nabla  \mathcal{L}(\tilde{\mathbf{M}})}_{\infty}\leq 2\smax^2a,$$
since for all $(i,j)\in\nint{1}{n}\times\nint{1}{p}$ $|\tilde{\mathbf{M}}_{ij} - \TX_{ij}|\leq 2a$ and $g_j''(\tilde{\mathbf{M}}_{ij})\leq \smax^2$. The last three inequalities plugged in \eqref{eq:th2-subres} give
\begin{equation*}
\begin{aligned}
& \frac{\smin^2}{2}\norm{\mathcal{P}_{\Omega}(\fu{\Delta \alp})}_F^2 && \leq \lambda_S\left(\norm{\Talpha}_{1}-\norm{\hat{\alp}}_{1}\right) +\umax\norm{\Delta\alp}_{1}\left\lbrace\norm{\nabla  \mathcal{L}(\TX)}_{\infty}+2\smax^2a\right\rbrace.
\end{aligned}
\end{equation*}
The triangular inequality gives
\begin{equation*}
\begin{aligned}
& \frac{\smin^2}{2}\norm{\mathcal{P}_{\Omega}(\fu{\Delta \alp})}_F^2 &&  \leq \left\lbrace\umax\left(\norm{\nabla  \mathcal{L}(\TX)}_{\infty} + 2\smax^2a\right)+\lambda_S\right\rbrace\norm{\Talpha}_{1}  \\
& && +\left\lbrace\umax\left(\norm{\nabla  \mathcal{L}(\TX)}_{\infty} + 2\smax^2a\right)-\lambda_S\right\rbrace\norm{\hat{\alp}}_{1}.
\end{aligned}
\end{equation*}
Then, the assumption $\lambda_S\geq 2\umax\left(\norm{\diff{\TX}}_{\infty}+ 2\smax^2(1+\nu)a\right)$ gives 
$$\norm{\mathcal{P}_{\Omega}(\fu{\Delta \alp})}_F^2 \leq \frac{3\lambda_S}{\smin^2}\norm{\Talpha}_{1}.$$
Plugged into \eqref{eq:upper-bound-alpha-subres}, this last inequality implies that with probability at least $1-8(n+p)^{-1}$
\begin{equation}
\label{eq:upper-bound-alpha-subres-2}
\EE\norm{\mathcal{P}_{\Omega}(\fu{\Delta\alp})}_F^2\leq \frac{3\lambda_S}{\smin^2}\norm{\Talpha}_{1} + 64\nu a\norm{\Talpha}_{1}\umax\pe{\norm{\Sigma_R}_{\infty}} + 3072a^2p^{-1}.
\end{equation}
Combining \eqref{eq:ineq-alpha-l1} and \eqref{eq:upper-bound-alpha-subres-2} gives the result.
\subsection{Proof of \Cref{Lemma:L-nuc}}
\label{Lemma:L-nuc-proof}

Using \eqref{eq:opt-cond} for $L=\TL$ and $\alp = \Talpha$ we obtain 
\begin{equation*}
\label{eq:subdiff-L}
\pscal{\diff{\hat{\mathbf{M}}}}{\Delta \Prm+\sum_{k=1}^q(\Delta \alp_k)\mathbf{X}(k)}+\lambda_L\pscal{\hat f_L}{\Delta \Prm}+\lambda_S\pscal{\hat f_{\alp}}{\Delta \alp}\geq 0.
\end{equation*}
Then, the convexity of $\lnuc$ and $\lone$ imply that
\begin{equation*}
\begin{aligned}
& \norm{\TL}_{\star}\geq \norm{\hat{\Prm}}_{\star}+\pscal{\partial\norm{\hat{\Prm}}_{\star}}{\Delta \Prm},\\
& \norm{\Talpha}_{1}\geq \norm{\hat \alp}_{\star}+\pscal{\partial\norm{\hat \alp}_{1}}{\Delta \alp}.
\end{aligned}
\end{equation*}
The last three inequalities yield
\begin{multline*}
\lambda_L\left(\norm{\hat{\Prm}}_{\star}-\norm{\TL}_{\star}\right) + \lambda_S\left(\norm{\hat \alp}_{1}-\norm{\Talpha}_{1}\right) \leq \pscal{\diff{\hat{\mathbf{M}}}}{\Delta \Prm}\\
+ \pscal{\diff{\hat{\mathbf{M}}}}{\sum_{k=1}^q(\Delta \alp_k)\mathbf{X}(k)}\\
\leq \norm{\diff{\hat{\mathbf{M}}}}\norm{\Delta \Prm}_{\star} + \umax\norm{\diff{\hat{\mathbf{M}}}}_{\infty}\norm{\Delta \alp}_{1}.
\end{multline*}
Using the conditions $$\lambda_L\geq 2\norm{\diff{\TX}},\quad\lambda_S\geq 2\umax\left\lbrace\norm{\diff{\TX}}_{\infty}+ 2\smax^2(1+\nu)a\right\rbrace,$$  
we get
\begin{multline*}
\lambda_L\left(\norm{P_{\TL}^{\perp}(\Delta \Prm)}_{\star}-\norm{P_{\TL}(\Delta \Prm)}_{\star}\right) + \lambda_S\left(\norm{\hat \alp}_{1}-\norm{\Talpha}_{1}\right)\leq\\
 \frac{\lambda_L}{2}\left(\norm{P_{\TL}^{\perp}(\Delta \Prm)}_{\star}+\norm{P_{\TL}(\Delta \Prm)}_{\star}\right) + \frac{\lambda_S}{2}\norm{\Delta \alp}_{1},
\end{multline*}
which implies
$$\norm{P_{\TL}^{\perp}(\Delta \Prm)}_{\star} \leq 3\norm{P_{\TL}(\Delta \Prm)}_{\star} + 3\lambda_S/\lambda_L\norm{\Talpha}_{1}.$$
Now, using $$\norm{\Delta \Prm}_{\star}\leq \norm{P_{\TL}^{\perp}(\Delta \Prm)}_{\star}+\norm{P_{\TL}(\Delta \Prm)}_{\star},\quad \norm{P_{\TL}(\Delta \Prm)}_F\leq \norm{\Delta \Prm}_F$$ and $\operatorname{rank}(P_{{\TL}}(\Delta \Prm))\leq 2 r$, we get
$$\norm{\Delta \Prm}_{\star} \leq \sqrt{32r}\norm{\Delta \Prm}_F + 3\lambda_S/\lambda_L\norm{\Talpha}_{1}.$$
This completes the proof of \Cref{Lemma:L-nuc}.

\subsection{Proof of \Cref{Lemma:rsc}}
\label{Lemma:rsc-proof}

\paragraph{Proof of (i):} Recall $$\mathsf{D}_{\alp} = 8\nu d_1\umax\pe{\norm{\Sigma_R}_{\infty}}+768p^{-1}$$ and
$$\tilde{\mathcal{A}}(d_1) = \left\lbrace \alp\in\R^q :\quad \norm{\alp}_{\infty}\leq 1;\quad \norm{\alp}_{1}\leq d_1;\quad  \EE\norm{\mathcal{P}_{\Omega}(\fu{\alp})}_F^2\geq \frac{18\log(n+p)}{\pi\log(6/5)}\right\rbrace.
$$
 We will show that the probability of the following event is small:
$$\mathcal{B} = \left\lbrace \exists \alp\in\tilde{\mathcal{A}}(d_1) \text{ such that } \left|\norm{\mathcal{P}_{\Omega}(\fu{\alp} )}_F^2 -   \EE\norm{\mathcal{P}_{\Omega}(\fu{\alp})}_F^2\right|>  \frac{1}{2} \EE\norm{\mathcal{P}_{\Omega}(\fu{\alp})}_F^2+ \mathsf{D}_{\alp}\right\rbrace.$$
Indeed, $\mathcal{B}$ contains the complement of the event we are interested in. We use a peeling argument to upper bound the probability of event $\mathcal{B}$. Let $\nu = \frac{18\log(n+p)}{\pi\log(6/5)}$ and $\eta = 6/5$. For $l\in\mathbb{N}$ set 
$$\mathcal{S}_l = \left\lbrace \alp\in\tilde{\mathcal{A}}(d_1):\quad \eta^{l-1}\nu\leq  \EE\norm{\mathcal{P}_{\Omega}(\fu{\alp})}_F^2 \leq \eta^l\nu \right\rbrace.$$ 
Under the event $\mathcal{B}$, there exists $l\geq 1$ and $\alp \in \tilde{\mathcal{A}}(d_1)\cap S_l$ such that 
\begin{equation}
\label{eq:peel}
\begin{aligned}
 &\left|\norm{\mathcal{P}_{\Omega}(\fu{\alp})}^2 -   \EE\norm{\mathcal{P}_{\Omega}(\fu{\alp})}_F^2\right|&&> \frac{1}{2} \EE\norm{\mathcal{P}_{\Omega}(\fu{\alp})}_F^2+ \mathsf{D}_{\alp}
\\
& && > \frac{1}{2}\eta^{l-1}\nu + \mathsf{D}_{\alp}\\
& && = \frac{5}{12}\eta^l\nu + \mathsf{D}_{\alp}.
\end{aligned}
\end{equation}
For $T>\nu$, consider the set of vectors
$$\tilde{\mathcal{A}}(d_1, T) = \left\lbrace \alp \in \tilde{\mathcal{A}}(d_1): \EE\norm{\mathcal{P}_{\Omega}(\fu{\alp})}_F^2\leq T\right \rbrace $$
and the event
$$\mathcal{B}_l = \left\lbrace \exists \alp\in \tilde{\mathcal{A}}(d_1, \eta^l\nu): \left|\norm{\mathcal{P}_{\Omega}(\fu{\alp})}_F^2  -   \EE\norm{\mathcal{P}_{\Omega}(\fu{\alp})}_F^2\right|>\frac{5}{12}\eta^l\nu + \mathsf{D}_{\alp} \right\rbrace.$$
If $\mathcal{B}$ holds, then \eqref{eq:peel} implies that $\mathcal{B}_l$ holds for some $l\leq 1$. Therefore ,$\mathcal{B}\subset \cup_{l=1}^{+\infty} \mathcal{B}_l$, and it is enough to estimate the probability of the events $\mathcal{B}_l$ and then apply the union bound. Such an estimation is given in the following Lemma, adapted from Lemma 10 in \cite{klopp:hal-01111757}. 
\begin{Lemma}
\label{Lemma:ZT-concentration}
Define
$Z_T = {\sup}_{\alp\in\tilde{\mathcal{A}}(d_1, T)}\left|\norm{\mathcal{P}_{\Omega}(\fu{\alp})}_F^2  -   \EE\norm{\mathcal{P}_{\Omega}(\fu{\alp})}_F^2\right|.$ Then,
$$\prob{Z_T \geq \mathsf{D}_{\alp}+\frac{5}{12}T}\leq 4\e^{-\pi T/18}.$$
\end{Lemma}
\Cref{Lemma:ZT-concentration} gives that $\prob{\mathcal{B}_l}\leq 4\exp(-\pi\eta^l\nu/18)$. Applying the union bound we obtain
\begin{equation*}
\begin{aligned}
&\prob{\mathcal{B}} && \leq \sum_{l=1}^{\infty}\prob{\mathcal{B}_l}\\
& && \leq 4\sum_{l=1}^{\infty}\exp(-\pi\eta^l\nu/18)\\
& && \leq 4\sum_{l=1}^{\infty}\exp(-\pi\log(\eta)l\nu/18),
\end{aligned}
\end{equation*} 
where we used $e^x\geq x$. Finally, for $\nu = \frac{18\log(n+p)}{\pi\log(6/5)}$ we obtain
$$\prob{\mathcal{B}}\leq \frac{4\exp(-\pi\nu\log(\eta)/18)}{1-\exp(-\pi\nu\log(\eta)/18)}\leq \frac{4\exp(-\log(n+p))}{1-\exp(-\log(n+p))}\leq \frac{8}{n+p}, $$
since $d-1\geq (n+p)/2$, which concludes the proof of (i).

\paragraph{Proof of (ii):} The proof is very similar to that of (i); we recycle some of the notations for simplicity. Recall
$$\mathsf{D}_X=\frac{112\rho}{\pi} \pe{\norm{\Sigma_R}}^2 + 8\nu\varepsilon \pe{\norm{\Sigma_R}}+8\nu d_1\umax\pe{\norm{\Sigma_R}_{\infty}} +d_{\Pi}+768p^{-1}.$$
Let\begin{multline*}
\mathcal{B} = \Big\lbrace \exists (\Prm,\alp)\in\mathcal{C}(d_1,d_{\Pi},\rho,\varepsilon);\\
 \left|\norm{\mathcal{P}_{\Omega}(\Prm+\fu{\alp})}_F^2 -  \EE\norm{\mathcal{P}_{\Omega}(\Prm+\fu{\alp})}_F^2\right|>  \frac{1}{2}\EE\norm{\mathcal{P}_{\Omega}(\Prm+\fu{\alp})}_F^2+ \mathsf{D}_X\Big\rbrace,
\end{multline*}
 $\nu = \frac{72\log(n+p)}{\pi\log(6/5)}$, $\eta = \frac{6}{5}$ and for $l\in\mathbb{N}$ $$\mathcal{S}_l = \left\lbrace (\Prm,\alp)\in\mathcal{C}(d_1,d_{\Pi},\rho,\varepsilon):\quad \eta^{l-1}\nu\leq \EE\norm{\mathcal{P}_{\Omega}(\Prm+\fu{\alp})}_F^2 \leq \eta^l\nu \right\rbrace.$$
As before, if $\mathcal{B}$ holds, then there exist $l\geq 2$ and $(\Prm,\alp)\in\mathcal{C}(d_1,d_{\Pi},\rho,\varepsilon)\cap S_l$ such that
\begin{equation}
\label{eq:peel2}
\begin{aligned}
 &\left|\norm{\mathcal{P}_{\Omega}(\Prm+\fu{\alp})}_F^2 -  \EE\norm{\mathcal{P}_{\Omega}(\Prm+\fu{\alp})}_F^2\right|&&> \frac{5}{12}\eta^l\nu + \mathsf{D}_X.
\end{aligned}
\end{equation}
For $T>\nu$, consider the set $\tilde{\mathcal{C}}(T) = \left\lbrace (\Prm,\alp) \in \mathcal{C}(d_1,d_{\Pi},\rho,\varepsilon): \EE\norm{\mathcal{P}_{\Omega}(\Prm+\fu{\alp})}_F^2\leq T\right \rbrace $, and the event
$$\mathcal{B}_l = \left\lbrace \exists (\Prm,\alp)\in \tilde{\mathcal{C}}(\eta^l\nu):\quad \left|\norm{\mathcal{P}_{\Omega}(\Prm+\fu{\alp})}_F^2 -  \EE\norm{\mathcal{P}_{\Omega}(\Prm+\fu{\alp})}_F^2\right|>\frac{5}{12}\eta^l\nu + \mathsf{D}_X \right\rbrace.$$
Then, \eqref{eq:peel2} implies that $\mathcal{B}_l$ holds and $\mathcal{B}\subset \cup_{l=1}^{+\infty} \mathcal{B}_l$. Thus, we estimate in \Cref{Lemma:ZT-concentration2} the probability of the events $\mathcal{B}_l$, and then apply the union bound. 
\begin{Lemma}
\label{Lemma:ZT-concentration2}
Let $W_T = {\sup}_{(\Prm,\alp)\in\tilde{\mathcal{C}}(T)}\left|\norm{\mathcal{P}_{\Omega}(\Prm+\fu{\alp})}_F^2 -  \EE\norm{\mathcal{P}_{\Omega}(\Prm+\fu{\alp})}_F^2\right|.$
$$\prob{W_T \geq \mathsf{D}_X+\frac{5}{12}T}\leq 4\e^{-\pi T/72}.$$
\end{Lemma}

\Cref{Lemma:ZT-concentration2} gives that $\prob{\mathcal{B}_l}\leq 4\exp(-\pi\eta^l\nu/72)$. Applying the union bound we obtain
\begin{equation*}
\begin{aligned}
&\prob{\mathcal{B}} && \leq \sum_{l=1}^{\infty}\prob{\mathcal{B}_l}\\
& && \leq 4\sum_{l=1}^{\infty}\exp(-\pi\eta^l\nu/72)\\
& && \leq 4\sum_{l=1}^{\infty}\exp(-\pi\log(\eta)l\nu/72),
\end{aligned}
\end{equation*} 
where we used $\e^x\geq x$. Finally, for $\nu = \frac{72\log(n+p)}{\pi\log(6/5)}$ we obtain
$$\prob{\mathcal{B}}\leq \frac{4\exp(-\pi\nu\log(\eta)/72)}{1-\exp(-\pi\nu\log(\eta)/72)}\leq \frac{4\exp(-\log(n+p))}{1-\exp(-\log(n+p))}\leq 8(n+p)^{-1}, $$
since $n+p-1\geq (n+p)/2$, which concludes the proof of (ii).

%% file: append.tex

\section{Proof of Theorem~\ref{prop:main}}
To prove the theorem, we first lower bound on the progress made by the algorithm
at the two blocks between the iterations.
With a slight abuse of notations, in the following we shall denote
the iterates without the bracket in the superscripts, e.g., we denote
$\alp^{(t)}, \Prm^{(t)}, R^{(t)}$ by $\alp^t, \Prm^t, R^t$, respectively, to simplify
our discussions.
 
For the first block on $\alp$, 
in Section~\ref{sec:1stblock} we show that
\beq
\label{eq:1stblock}
{F}( \alp^t, \Prm^{t-1}, R^{t-1} ) \leq {F}( \alp^{t-1} , \Prm^{t-1}, R^{t-1} ) - \frac{\gamma}{2}
\frac{\big( g_{\alp} ( \alp^{t-1}, \Prm^{t-1} ; Q^{t-1} ) \big)^2}{ (2 Q^{t-1})^2  } \eqs,
\eeq
where $Q^{t-1} \eqdef  \lambda_S^{-1} F( \alp^{t-1}, \Prm^{t-1}, R^{t-1} )$
as defined in the main paper and
\beq
g_{\alp} ( \alp^{t-1}, \Prm^{t-1} ; Q^{t-1} )  \eqdef 
\langle \grd_{\alp} {\cal L} ( \alp^{t-1}, \Prm^{t-1} ), \alp^{t-1} - \hat{\alp}^{t-1} \rangle 
+ \lambda_S ( \| \alp^{t-1} \|_1 - \| \hat{\alp}^{t-1} \|_1 ) \eqs,
\eeq
such that
\beq
\hat{\alp}^{t-1} \eqdef \argmin_{ \alp }~\big( \langle 
\grd_{\alp} {\cal L} ( \alp^{t-1}, \Prm^{t-1} ), \alp \rangle + \lambda_S \| \alp \|_1 \big)~~{\rm s.t.}~~
\| \alp \|_1 \leq Q^{t-1} \eqs.
\eeq

For the second block on $(\Prm, R)$, Section~\ref{sec:2ndblock} shows that
\beq\label{eq:2ndblock}
{F}( \alp^t, \Prm^t , R^t ) \leq {F}( \alp^t, \Prm^{t-1}, R^{t-1} ) - \frac{(g_{\Prm} ( \alp^t, \Prm^{t-1}, R^{t-1} ; R_{\sf UB}^t ))^2}{\max\{ 2 R_{\sf UB}^t ( \lambda_L + M^t), 8 \sigma_{\Prm} (R_{\sf UB}^t)^2 \}} \eqs,
\eeq
where $M^t \eqdef \| \grd_{\Prm} ( \alp^t,  \Prm^{t-1}) \|_2$ and we recall that $R_{\sf UB}^t \eqdef \lambda_L^{-1} F( \alp^t, \Prm^{t-1}, R^{t-1} )$
and we have defined
\beq \label{eq:g_fct_real}
g_{\Prm} ( \alp^t, \Prm^{t-1}, R^{t-1} ; R_{\sf UB}^t ) \eqdef 
\langle \Prm^{t-1} - \hat{\Prm}^t , \grd_{\Prm} {\cal L}(  \alp^t, \Prm^{t-1} ) \rangle 
+ \lambda_L ( {R}^{t-1} - \hat{R}^t  ) \eqs.
\eeq 
Moreover, Section~\ref{sec:2ndblock} shows that
\beq \label{eq:2ndblock_a}
{F}( \alp^t, \Prm^t , R^t ) -  {F}( \alp^t, \Prm^{t-1}, R^{t-1} ) \leq - \frac{\sigma_{\Prm}}{2} \| 
{\cal P}_\Omega( \Prm^t - \Prm^{t-1} ) \|_F^2 \eqs.
\eeq

\textbf{Statement (i)}. 
The above results show that the objective values for the iterates
produced by the MCGD method are non-increasing, \ie 
\beq
{F}( \alp^t, \Prm^{t}, R^{t} ) \leq {F}( \alp^t, \Prm^{t-1}, R^{t-1} ) \leq {F}( \alp^{t-1}, \Prm^{t-1}, R^{t-1} )
\eeq 
Now, 
consider the time varying part in the quantity $C(t)$ [cf.~\eqref{eq:Ct}] --- 
$Q^t$, $R_{\sf UB}^t$, $M^t$. The first two quantities are defined from the 
objective values and are thus bounded by 
$\lambda_S^{-1}F( \alp^0, \Prm^0, R^0)$, 
$\lambda_L^{-1}F( \alp^0, \Prm^0, R^0)$, respectively. 
Moreover, from the monotonicity of 
$F(\alp^t,\Prm^{t-1}, R^{t-1})$, we have
$\lambda_L \| \Prm^{t-1} \|_\star + \lambda_S \| \alp^t \|_1 \leq F( \alp^t, \Prm^{t-1}, R^{t-1} ) \leq F( \alp^0, \Prm^0, R^0 )$ for all $t \geq 1$.
As the gradient $\grd_{\Prm}{\cal L} ( \alp, \Prm )$ 
is bounded whenever $\alp, \Prm$ are bounded, we conclude that $M^t$ is bounded, e.g., 
$M^t \leq \bar{M}$. 
Finally, this shows for all $t \geq 1$ that
\beq
C(t) \leq \overline{C} \eqdef 
\max \Big\{ \frac{24 (Q^0)^2}{\gamma}, \frac{24 \hat{\sigma}_{\Prm}^2 (Q^0)^2}{\sigma_{\Prm}} + \max\{ 6 R_{\sf UB}^0 ( \lambda_L + \bar{M}), 24 \sigma_{\Prm} (R_{\sf UB}^0)^2 \} \Big\}
\eqs.
\eeq


\textbf{Statement (ii)}. To characterize the convergence rate of the MCGD method, 
let us consider the Lyapunov function, $g^t ( Q^t, R_{\sf UB}^t )$, defined as:
\beq
g^{t} ( Q^t, R_{\sf UB}^t ) \eqdef g_{\alp} ( \alp^{t}, \Prm^{t-1} ; Q^t ) + g_{\Prm} ( \alp^{t}, \Prm^{t-1}, R^{t-1} ; R_{\sf UB}^t ) \eqs.
\eeq
Note that as the loss function ${\cal L}( \alp, \Prm)$ is convex
and $\| \hat{\alp} \|_1 \leq Q^t$, $\| \hat{\Prm} \|_\star \leq R_{\sf UB}^t$, 
it is possible to lower bound 
$g^t ( Q^t, R_{\sf UB}^t )$ by:
\beq
g^t ( Q^t, R_{\sf UB}^t ) \geq F( \alp^t, \Prm^{t-1}, R^{t-1} ) - F( \hat{\alp}, \hat{\Prm}, \hat{R} ) \eqs.
\eeq

Furthermore, we can 
obtain an upper bound to $g^t ( Q^t, R_{\sf UB}^t )$ in terms of the objective values:
\beq
\begin{split}
& g_{\alp} ( \alp^{t}, \Prm^{t-1} ; Q^t ) = \max_{ \| \alp \| \leq Q^t }
\langle \grd_{\alp} {\cal L} ( \alp^{t}, \Prm^{t-1} ), \alp^{t} - {\alp} \rangle 
+ \lambda_S ( \| \alp^{t} \|_1 - \| {\alp} \|_1 ) \\
& = \max_{ \| \alp \|_1 \leq Q^t } \langle \grd_{\alp} {\cal L} ( \alp^{t}, \Prm^{t} ), \alp^{t} - {\alp} \rangle 
+ \langle \grd_{\alp} {\cal L} ( \alp^{t}, \Prm^{t-1} ) - \grd_{\alp} {\cal L} ( \alp^{t}, \Prm^{t} ), \alp^{t} - {\alp} \rangle \\
& \hspace{2cm} + \lambda_S \big( \| \alp^t \|_1 - \| \alp \|_1 \big) \\[.1cm]
& \leq \max_{ \| \alp \|_1 \leq Q^t } \langle \grd_{\alp} {\cal L} ( \alp^{t}, \Prm^{t} ), \alp^{t} - {\alp} \rangle + \lambda_S \big( \| \alp^t \|_1 - \| \alp \|_1 \big) \\
& \hspace{2cm} + \| \grd_{\alp} {\cal L} ( \alp^{t}, \Prm^{t-1} ) - \grd_{\alp} {\cal L} ( \alp^{t}, \Prm^{t} ) \|_2 \| \alp^{t} - {\alp} \|_2 \\[.1cm]
& \leq g_{\alp} ( \alp^{t}, \Prm^{t} ; Q^t ) + 2 \hat{\sigma}_{\Prm} Q^t \| {\cal P}_\Omega( \Prm^{t-1}  - \Prm^{t} ) \|_F \eqs.
\end{split}
\eeq
Consequently, we have
\beq \notag
\begin{split}
& \big( g^{t} ( Q^t, R_{\sf UB}^t ) \big)^2 \\
& \leq 3 \big( 
(g_{\Prm} ( \alp^{t}, \Prm^{t-1}, R^{t-1} ; R_{\sf UB}^t ))^2 +
(g_{\alp} ( \alp^{t}, \Prm^{t} ; Q^t ))^2 +  4
\hat{\sigma}_{\Prm}^2 (Q^t)^2 \| {\cal P}_\Omega( \Prm^{t-1}  - \Prm^{t} ) \|_F^2 \big) \\
& \leq 3 \big( C_1^t \big( F( \alp^t, \Prm^t, R^t ) - F( \alp^{t+1}, \Prm^t, R^t ) \big) + C_2^t \big( 
F( \alp^t, \Prm^{t-1} , R^{t-1} ) - F( \alp^t, \Prm^t, R^t) \big) \big)
\end{split}
\eeq
where
\beq
C_1^t \eqdef \frac{ 8(Q^t)^2}{ \gamma },~~C_2^t = \frac{ 8 \hat{\sigma}_{\Prm}^2 ( Q^t)^2 }
{ \sigma_{\Prm} } + \max\{ 2 R_{\sf UB}^t ( \lambda_L + M), 8 \sigma_{\Prm} (R_{\sf UB}^t)^2 \}
\eeq
Observe that $C(t)$ is defined by
$C(t) = 3 \max\{ C_1^t, C_2^t \}$ as the upper bound of the above
constants, we get
\beq
\big( g^{t} ( Q^t, R_{\sf UB}^t ) \big)^2 \leq C(t) \big( F( \alp^t, \Prm^{t-1}, R^{t-1} ) - F( \alp^{t+1}, \Prm^t, R^t ) \big) \eqs.
\eeq
Using the shorthand notation $\Delta^t \eqdef F( \alp^t, \Prm^{t-1}, R^{t-1} ) - F( \hat{\alp}, \hat{\Prm}, \hat{R} )$ and notice that $\big( g^{t} ( Q^t, R_{\sf UB}^t ) \big)^2 \geq (\Delta^t)^2$, we arrive at the following inequality:
\beq
\Delta^{t+1} \leq \Delta^t - \frac{1}{C(t)} ( \Delta^t )^2
\eeq
Applying Lemma~\ref{lem:n} in Section~\ref{sec:n_lemma}, we can show that
\beq
\Delta^{t+1} \leq \frac{1}{ (\Delta^1)^{-1} +  \sum_{i=1}^t \frac{1}{C(i)} } \eqs,
\eeq
Note that $\Delta^1 \leq \tilde{\Delta}^0 \eqdef F( \alp^0, \Prm^0, R^0 ) - F(\hat{\alp}, \hat{\Prm}, \hat{R})$, we have
\beq
\Delta^{t+1} \leq
\frac{1}{ (\tilde{\Delta}^0)^{-1} +  \sum_{i=1}^t \frac{1}{C(i)} } \leq \frac{ 1 }{ (\tilde{\Delta}^0)^{-1}  +  t \overline{C}(t) },~\forall~t \geq 0 \eqs.
\eeq
The proof is concluded by the straightforward inequality $F_0( \alp^{t+1} , \Prm^{t+1} ) - F_0(\hat{\alp} , \hat{\Prm}) \leq F( \alp^{t+1}, \Prm^{t+1} , R^{t+1} ) - F(\hat{\alp} , \hat{\Prm}, \hat{R} ) \leq \Delta^{t+1}$. 

\textbf{Comment on $\lim_{t \rightarrow \infty} C(t)$}. Since both 
$F( \alp^t, \Prm^t, R^t )$ and $F( \alp^t, \Prm^{t-1}, R^{t-1})$ converge to 
$F^\star \eqdef F( \hat{\alp}, \hat{\Prm}, \hat{R})$, \ie the optimal objective value. 
It is clear that $Q^t \rightarrow \hat{Q} \eqdef \lambda_S^{-1} F^\star$ and $R_{\sf UB}^t
\rightarrow \hat{R}_{\sf UB} \eqdef \lambda_L^{-1} F^\star$ as well. 
Furthermore, by continuity of the gradient, 
we have $M^t \rightarrow \| \grd_{\Prm} {\cal L} ( \hat{\alp}, \hat{\Prm} ) \|_2$. 
This shows that the limit $C^\star = \lim_{t \rightarrow \infty} C(t)$ exists. 

To obtain a computable bound for $C^\star$, 
note that $(\hat{\alp}, \hat{\Prm})$ is also an optimal solution 
to \eqref{eq:opt} and the optimality condition shows that
\beq
{\bm 0} \in \grd_{\Prm} {\cal L} ( \hat{\alp}, \hat{\Prm} ) + \lambda_L \partial \| \hat{\Prm} \|_\star
\eeq
By \cite[P.~41]{watson1992characterization}, 
we know that $\partial \| \hat{\Prm} \|_\star = \{ {\bm U}_1 {\bm V}_1^\top + {\bm W} ~:~ \| {\bm W} \|_2 \leq 1,~{\bm U}_1^\top {\bm W} = {\bm 0},~{\bm W} {\bm V}_1 = {\bm 0} \}$
such that ${\bm U}_1 \in \RR^{m_1 \times r}, {\bm V}_1 \in \RR^{m_2 \times r}$ 
are the left/right singular vectors of $\hat{\Prm}$
corresponding the $r \eqdef {\rm rank}( \hat{\Prm})$ non-zero singular values
of $\hat{\Prm}$. Importantly, this implies that
$\| \grd_{\Prm} {\cal L} ( \hat{\alp}, \hat{\Prm} ) \|_2 \leq 2 \lambda_L$ and 
\beq
C^\star \leq \overline{C}^\star \eqdef
\max \Big\{ \frac{24 (\hat{Q})^2}{\gamma}, \frac{24 \hat{\sigma}_{\Prm}^2 (\hat{Q})^2}{\sigma_{\Prm}} + \max\{ 18 \hat{R}_{\sf UB} \lambda_L, 24 \sigma_{\Prm} (\hat{R}_{\sf UB})^2 \} \Big\} \eqs.
\eeq

\subsection{Proof of Eq.~\eqref{eq:1stblock}} \label{sec:1stblock}
Suppose $\alp^t$ is obtained by the proximal update in \eqref{eq:prox}, we observe that
\beq
\begin{split}
F( \alp^t, \Prm^{t-1}, R^{t-1} ) & \leq F( \alp^{t-1}, \Prm^{t-1}, R^{t-1} ) + \langle \grd_{\alp} {\cal L}( \alp^{t-1}, \Prm^{t-1} ) , \alp^t - \alp^{t-1} \rangle \\
& \hspace{.4cm} + \frac{ \sigma_{\alp} }{2} \| \alp^t - \alp^{t-1} \|_2^2 + \lambda_S \big( \| \alp^t \|_1 - \| \alp^{t-1} \|_1 \big) \eqs.
\end{split}
\eeq
On the other hand, when 
$\alp^t$ is obtained by the exact minimization in \eqref{eq:exact_pg}, denoted by 
$\alp_{\sf exact}^t$ to avoid confusion, 
we have $F( \alp_{\sf exact}^t , \Prm^{t-1}, R^{t-1} ) \leq F( \alp^t, \Prm^{t-1}, R^{t-1} )$
since the latter is an exact minimizer. 
Thus, $F( \alp_{\sf exact}^t , \Prm^{t-1}, R^{t-1} ) $ is upper bounded by the right hand side 
in the above inequality. 

Using the property of the proximal operator, it can be shown that
\beq
\alp^t \in \argmin_{ \alp } \Big( \langle \grd_{\alp} {\cal L}( \alp^{t-1}, \Prm^{t-1} ), \alp - \alp^{t-1} \rangle + \frac{1}{2 \gamma} \| \alp - \alp^{t-1} \|_2^2 + \lambda_S (\| \alp \|_1 - \|\alp^{t-1} \|_1)  \Big) 
\eeq
Due to our choice of step size, we have $\sigma_{\alp} \leq 1 / \gamma$. Combining this with
the above inequality implies that
\beq
\begin{split}
F( \alp^t, \Prm^{t-1}, R^{t-1} ) & \leq F( \alp^{t-1}, \Prm^{t-1}, R^{t-1} ) + \langle \grd_{\alp} {\cal L}( \alp^{t-1}, \Prm^{t-1} ) , \alp - \alp^{t-1} \rangle \\
& \hspace{.4cm} + \frac{ 1 }{2 \gamma} \| \alp - \alp^{t-1} \|_2^2 + \lambda_S \big( \| \alp \|_1 - \| \alp^{t-1} \|_1 \big),~\forall~\alp \in \RR^K \eqs.
\end{split}
\eeq
Furthermore, for all $b \in \RR$ it holds that
\beq
\begin{split}
F( \alp^t, \Prm^{t-1}, R^{t-1} ) & \leq F( \alp^{t-1}, \Prm^{t-1}, R^{t-1} ) + b \langle \grd_{\alp} {\cal L}( \alp^{t-1}, \Prm^{t-1} ) , \hat{\alp}^{t-1} - \alp^{t-1} \rangle \\
& \hspace{.2cm} + \frac{ b^2 }{2 \gamma} \| \hat{\alp}^{t-1} - \alp^{t-1} \|_2^2 + \lambda_S \big( \| b \hat{\alp}^{t-1} + (1-b) \alp^{t-1} \|_1 - \| \alp^{t-1} \|_1 \big) \\
& \leq F( \alp^{t-1}, \Prm^{t-1}, R^{t-1} ) + b \langle \grd_{\alp} {\cal L}( \alp^{t-1}, \Prm^{t-1} ) , \hat{\alp}^{t-1} - \alp^{t-1} \rangle \\
& \hspace{.2cm} + \frac{ b^2 }{2 \gamma} \| \hat{\alp}^{t-1} - \alp^{t-1} \|_2^2 + b \lambda_S \big( \| \hat{\alp}^{t-1}  \|_1 - \| \alp^{t-1} \|_1 \big) \eqs,
\end{split}
\eeq
where we have limited our search space from $\alp \in \RR^K$ to $\alp = b \hat{\alp}^t + (1-b) \alp^{t-1}$ for $b \in \RR$. 
Minimizing the right hand side of the above with respect to $b$ yields
\beq
\begin{split}
& F( \alp^t, \Prm^{t-1}, R^{t-1} ) - 
F( \alp^{t-1}, \Prm^{t-1}, R^{t-1} ) \\
& \leq - \frac{ \gamma }{2} \frac{ (g_{\alp} ( \alp^{t-1}, \Prm^{t-1} ; Q^{t-1} ))^2 }{ \| \hat{\alp}^{t-1} - \alp^{t-1} \|_2^2 }
\leq - \frac{ \gamma }{2} \frac{ (g_{\alp} ( \alp^{t-1}, \Prm^{t-1} ; Q^{t-1} ))^2 }{ (2 Q^{t-1})^2 } \eqs,
\end{split}
\eeq
where we have used 
$\| \hat{\alp}^{t-1} - \alp^{t-1} \|_2^2 \leq (2 Q^{t-1})^2$ in the last inequality.

\subsection{Proof of Eq.~\eqref{eq:2ndblock} and \eqref{eq:2ndblock_a}} \label{sec:2ndblock}
Let us observe that
\beq
\begin{split}
F( \alp^t, \Prm^t, R^t ) & = F( \alp^t, \Prm^{t-1}, R^{t-1} ) - \beta_t g_{\Prm} ( \alp^t, \Prm^{t-1}, R^{t-1}; R_{\sf UB}^t ) \\
& \hspace{.5cm} + \frac{\beta_t^2}{2} \left( \begin{array}{c}
{\rm vec}( \hat{\Prm}^t - \Prm^{t-1} ) \\
\hat{R}^t - R^{t-1} 
\end{array} \right)^\top \grd_{ \Prm, R }^2 ( \bm{\xi} ) 
\left( \begin{array}{c}
{\rm vec}( \hat{\Prm}^t - \Prm^{t-1} ) \\
\hat{R}^t - R^{t-1} 
\end{array} \right) \eqs,
\end{split}
\eeq
where $\bm{\xi}$ is any point that lies on the line 
$[ ({\rm vec}( \Prm^{t-1} ); R^{t-1}), ({\rm vec}( \Prm^{t} ); R^{t}) ]$. 
From the property of $F$, we observe that
\beq
\grd_{ \Prm, R }^2 ( \bm{\xi} )
\preceq 
\left( 
\begin{array}{cc}
\sigma_{\Prm} {\rm Diag} ( {\cal P}_{\Omega} ({\bf J}) ) & {\bm 0} \\
{\bm 0} & {\bm 0} 
\end{array}
\right) \eqs,
\eeq
where ${\bf J}$ is the $m_1 \times m_2$ all-ones matrix. 
The above implies that
\beq
\begin{split}
F( \alp^t, \Prm^t, R^t ) & \leq F( \alp^t, \Prm^{t-1}, R^{t-1} ) - \beta_t g_{\Prm} ( \alp^t, \Prm^{t-1}, R^{t-1}; R_{\sf UB}^t ) \\
& \hspace{.5cm} + \frac{\beta_t^2 \sigma_{\Prm} }{2} \| {\cal P}_{\Omega} ( \hat{\Prm}^t - \Prm^{t-1} ) \|_F^2 \eqs.
\end{split}
\eeq
Recall that $\beta_t = \min\{ 1, g_{\Prm} ( \alp^t, \Prm^{t-1}, R^{t-1}; R_{\sf UB}^t ) / ( \sigma_{\Prm} 
\| {\cal P}_{\Omega} ( \hat{\Prm}^t - \Prm^{t-1} ) \|_F^2 ) \}$. 
If $g_{\Prm} ( \alp^t, \Prm^{t-1}, R^{t-1}; R_{\sf UB}^t ) \geq \sigma_{\Prm} 
\| {\cal P}_{\Omega} ( \hat{\Prm}^t - \Prm^{t-1} ) \|_F^2 $, then we choose $\beta_t = 1$
and observe:
\beq
\begin{split}
& F( \alp^t, \Prm^t, R^t ) - F( \alp^t, \Prm^{t-1}, R^{t-1} ) \\
& \leq  - \frac{1}{2} ~g_{\Prm} ( \alp^t, \Prm^{t-1}, R^{t-1}; R_{\sf UB}^t ) = - \frac{1}{2} \frac{ (g_{\Prm} ( \alp^t, \Prm^{t-1}, R^{t-1}; R_{\sf UB}^t ))^2 }{ g_{\Prm} ( \alp^t, \Prm^{t-1}, R^{t-1}; R_{\sf UB}^t ) }\\
& \leq -\frac{1}{2} \frac{(g_{\Prm} ( \alp^t, \Prm^{t-1}, R^{t-1}; R_{\sf UB}^t ))^2} { R_{\sf UB}^t ( \lambda_L + 2M^t) }  \eqs,
\end{split}
\eeq
where we have 
used the upper bound 
to $g_{\Prm} ( \alp^t, \Prm^{t-1}, R^{t-1}; R_{\sf UB}^t )$ as follows:
\beq \label{eq:inc_res1}
\begin{split}
g_{\Prm} ( \alp^t, \Prm^{t-1}, R^{t-1}; R_{\sf UB}^t ) & \leq \lambda_L R_{\sf UB}^t + \langle \Prm^{t-1} - \hat{\Prm}^t , \grd_{\Prm} {\cal L}(  \alp^t, \Prm^{t-1} )  \rangle \\
& \leq R_{\sf UB}^t  \big( \lambda_L + 2 M^t \big) \eqs,
\end{split}
\eeq
with $M^t \eqdef \| \grd_{\Prm} {\cal L}(  \alp^t, \Prm^{t-1} )  \|_2$ being
the spectral norm of the gradient. 

Otherwise, we choose $\beta_t = g_{\Prm} ( \alp^t, \Prm^{t-1}, R^{t-1}; R_{\sf UB}^t ) / ( \sigma_{\Prm} 
\| {\cal P}_{\Omega} ( \hat{\Prm}^t - \Prm^{t-1} ) \|_F^2 )$ and observe:
\beq \label{eq:inc_res2}
\begin{split}
& F( \alp^t, \Prm^t, R^t ) - F( \alp^t, \Prm^{t-1}, R^{t-1} ) \\
& \leq - \frac{1}{2} ~\frac{(g_{\Prm} ( \alp^t, \Prm^{t-1}, R^{t-1}; R_{\sf UB}^t ))^2}{\sigma_{\Prm} \| {\cal P}_{\Omega} ( \hat{\Prm}^t - \Prm^{t-1} ) \|_F^2 }
\leq -\frac{1}{2}~\frac{(g_{\Prm} ( \alp^t, \Prm^{t-1}, R^{t-1}; R_{\sf UB}^t ))^2}{  \sigma_{\Prm} (2R_{\sf UB}^t)^2}  \eqs,
\end{split}
\eeq
where we have used   
$\| {\cal P}_{\Omega} ( \hat{\Prm}^t - \Prm^{t-1} ) \|_F^2 \leq \| \hat{\Prm}^t - \Prm^{t-1} \|_F^2
\leq \| \hat{\Prm}^t - \Prm^{t-1} \|_\star^2 \leq (2 R_{\sf UB}^t)^2$.

To prove \eqref{eq:2ndblock_a}, we observe that
\beq
\| {\cal P}_\Omega (\Prm^t - \Prm^{t-1}) \|_F^2 = \beta_t^2 \| 
{\cal P}_\Omega ( \hat{\Prm}^t - \Prm^{t-1} ) \|_F^2 \eqs.
\eeq
If $\beta_t = 1$, then we have 
$g_{\Prm} ( \alp^t, \Prm^{t-1}, R^{t-1}; R_{\sf UB}^t ) \geq \sigma_{\Prm} 
\| {\cal P}_{\Omega} ( \hat{\Prm}^t - \Prm^{t-1} ) \|_F^2 $ and therefore we can upper bound
$\| {\cal P}_\Omega (\Prm^t - \Prm^{t-1}) \|_F^2$ by:
\beq
\frac{1}{ \sigma_{\Prm}} g_{\Prm} ( \alp^t, \Prm^{t-1}, R^{t-1}; R_{\sf UB}^t )
\leq \frac{2}{\sigma_{\Prm}} \Big( F( \alp^t, \Prm^{t-1}, R^{t-1} ) - F( \alp^t, \Prm^{t}, R^{t} ) \Big) 
\eeq
where the last inequality follows from \eqref{eq:inc_res1}. 
Otherwise, we choose $\beta_t = g_{\Prm} ( \alp^t, \Prm^{t-1}, R^{t-1}; R_{\sf UB}^t ) / 
\sigma_{\Prm} 
\| {\cal P}_{\Omega} ( \hat{\Prm}^t - \Prm^{t-1} ) \|_F^2$ and therefore,
\beq
\begin{split}
\| {\cal P}_\Omega (\Prm^t - \Prm^{t-1}) \|_F^2 & = \frac{1}{\sigma_{\Prm}} \frac{ (g_{\Prm} ( \alp^t, \Prm^{t-1}, R^{t-1}; R_{\sf UB}^t ))^2 }{\sigma_{\Prm} \| {\cal P}_{\Omega} ( \hat{\Prm}^t - \Prm^{t-1} ) \|_F^2} \\
& \leq \frac{2}{\sigma_{\Prm}} \Big( F( \alp^t, \Prm^{t-1}, R^{t-1} ) - F( \alp^t, \Prm^{t}, R^{t} ) \Big) \eqs,
\end{split}
\eeq
where the last inequality follows from \eqref{eq:inc_res2}. 

\subsection{Additional Lemma} \label{sec:n_lemma}
The following lemma is modified from \cite[Lemma 3.5]{beck2013convergence}.
\begin{Lemma} \label{lem:n}
Let $\{ A_k \}_{k \geq 1}$ be a non-negative sequence satisfying:
\beq
A_{k+1} \leq A_k - \gamma_k A_k^2,~k \geq 1 \eqs,
\eeq
where $\gamma_k$ is some positive number for all $k \geq 1$. Then,
\beq \label{eq:lem_fin}
A_{k+1} \leq \frac{ 1 }{ \frac{1}{A_1} + \sum_{i=1}^k \gamma_i },~k \geq 1 \eqs. 
\eeq
\end{Lemma}

\emph{Proof}: Consider the following chain of inequality:
\beq
\frac{1}{A_{k+1}} - \frac{1}{A_{k}} = \frac{ A_{k} - A_{k+1} }{ A_k A_{k+1} } 
\geq \gamma_k \frac{ A_k }{ A_{k+1} }  \geq \gamma_k \eqs,
\eeq
where the last inequality is due to the fact that $A_{k+1} \leq A_k$. 
Consequently, we have
\beq
\frac{1}{A_{k+1}} - \frac{1}{A_1} = \sum_{i=1}^k \Big( \frac{1}{A_{i+1}} - \frac{1}{A_i} \Big) 
\geq \sum_{i=1}^k \gamma_i \eqs.
\eeq
Reshuffling terms shows the desired result in \eqref{eq:lem_fin}. \hfill \textbf{Q.E.D.}

\section{Distributed MCGD Optimization}
Similar to the previous section, in the following we shall denote
the iterates without the bracket in the superscripts, e.g., we denote
$\alp^{(t)}, \Prm^{(t)}, R^{(t)}$ by $\alp^t, \Prm^t, R^t$, respectively, to simplify
our discussions.

Let us describe a distributed version of the MCGD method under a \emph{master-slave} 
architecture setting
where there exists $K$ workers and each of them is connected to a central server. 
Our goal is to offload the computation required by MCGD method to the workers, while 
protecting the privacy sensitive data owned by the workers. 
To describe our setting, the set of observed data ${\bf Y}_{ij},~(i,j) \in \Omega$ 
are stored
in $K$ different workers, where the $k$th worker holds ${\bf Y}_{ij}$ with
$(i,j) \in \Omega_k \subset \Omega$. 
Particularly, we have $\Omega = \Omega_1 \cup \cdots \cup \Omega_K$ with 
$\Omega_k \cap \Omega_{k'} = \emptyset$ for all $k \neq k'$. In this way, we can write
\beq
{\cal L} ( \alp, \Prm ) = \sum_{ (i,j) \in \Omega } \left\{ - {\bf Y}_{ij} {\bf M}_{ij} + g_j ( {\bf M}_{ij} )  \right\}
= \sum_{k=1}^K \underbrace{\sum_{ (i,j) \in \Omega_k } \left\{ - {\bf Y}_{ij} {\bf M}_{ij} + g_j ( {\bf M}_{ij} )  \right\}}_{\eqdef {\cal L}_k( \alp, \Prm )}
\eeq
such that the log-likelihood function can be decomposed as
${\cal L} ( \alp, \Prm ) \eqdef \sum_{k=1}^K {\cal L}_k ( \alp, \Prm )$.  
Moreover, notice that ${\cal L}_k ( \alp, {\cal P}_{\Omega_k}( \Prm ) ) = {\cal L}_k ( \alp, \Prm )$
since the $k$th local function is evaluated only on the entries in $\Omega_k$. 
For simplicity, we assume that computation can be done synchronously among the workers. 

We can implement the MCGD method in a distributed setting as follows. We focus
on the $t$th iteration where $\alp^{t-1}, \Prm^{t-1}, R^{t-1}$ have been previously computed
and worker $k$ now holds $\alp^{t-1}, {\cal P}_{\Omega_k} ( \Prm^{t-1} ), R^{t-1}$.

Firstly, the proximal update step of line~3 is replaced by a natural
distributed implementation
where the workers
compute and transmit the local gradients of the
log-likelihood function, $\grd_{\alp} {\cal L}_k ( \alp^{t-1}, {\cal P}_{\Omega_k} ( \Prm^{t-1} ) )$, 
to the master node;
the master node can then \emph{aggregate} the received local gradients 
to form the update in \eqref{eq:prox}, yielding $\alp^t$ which is then transmitted back to the 
workers.  

Secondly,  
the CG update of line~5 requires the top SVD of 
$\grd_{\Prm} {\cal L} (\alp^t,  \Prm^{t-1} )$ whose complexity is 
${\cal O}( | \Omega | \max\{ n,p \} \log ( 1/ \delta ) )$ using a centralized implementation,
where $\delta > 0$ is the desired accuracy of SVD. 
In a distributed setting, we can replace 
the step by a \emph{distributed power method} for offloading the complexity.  
Importantly, we observe that the top singular vectors of 
$\grd_{\Prm} {\cal L} (\alp^t,  \Prm^{t-1} )$ can be \emph{approximated} by 
the following power method recursions:
\algsetup{indent=0.75em}
\begin{algorithm}[H]
\caption{Distributed Power Method for MCGD.}\label{alg:dpm}
  \begin{algorithmic}[1]
  \STATE \textbf{Initialize:} initialization --- ${\bm u}(0) \sim {\cal N}( {\bm 0}, {\bm I} ) \in \RR^n$, and the parameter $P \in \ZZ$.
  \FOR {$p=1,2,\dots,P$}
  \STATE The central server sends the vector ${\bm u}(p-1)$ to workers. 
  \STATE For all $k$, worker $k$ computes the vector:
  \beq
  {\bm v}_k( p ) = \grd_{\Prm} {\cal L}_k (\alp^t, {\cal P}_{\Omega_k} ( \Prm^{t-1})  ) {\bm u}(p-1) 
  \eeq
  and transmit it to the central server.
  \STATE The central server forms the next iterate by ${\bm v}(p) = \sum_{k=1}^K {\bm v}_k ( p)$ and sends the vector ${\bm v}(p)$ to workers.
  \STATE For all $k$, worker $k$ computes the vector:
  \beq
  {\bm u}_k( p ) = \grd_{\Prm} {\cal L}_k (\alp^t, {\cal P}_{\Omega_k} ( \Prm^{t-1})  )^\top {\bm v}(p) 
  \eeq
  and transmit it to the central server.   
  \STATE The central server forms the next iterate by ${\bm u}(p) = \sum_{k=1}^K {\bm u}_k ( p)$.
\ENDFOR
\STATE At the central server, 
compute the top left and right singular vector as ${\bm u}_{(1)}^t = {\bm u}(P) / \| {\bm u}(P) \|$
and ${\bm v}_{(1)}^t = {\bm v}(P) / \| {\bm v}(P) \|$. 
\STATE \textbf{Return:} the top singular vectors ${\bm u}_{(1)}^t, {\bm v}_{(1)}^t$.
  \end{algorithmic}
\end{algorithm}
Line~4 and 5 in the above pseudo code implement the following power iterations:
\beq \label{eq:power_orig}
{\bm v}(p) = \grd_{\Prm} {\cal L} (\alp^t,  \Prm^{t-1} )
{\bm u}(p-1) = \sum_{k=1}^K \grd_{\Prm} {\cal L}_k (\alp^t,  {\cal P}_{\Omega_k} (\Prm^{t-1}) )
{\bm u}(p-1)
\eeq
\beq \label{eq:power_orig2}
{\bm u}(p) = \grd_{\Prm} {\cal L} (\alp^t,  \Prm^{t-1} )^\top
{\bm v}(p) = \sum_{k=1}^K \grd_{\Prm} {\cal L}_k (\alp^t,  {\cal P}_{\Omega_k} (\Prm^{t-1}) )^\top
{\bm v}(p) \eqs,
\eeq
where we have exploited the decomposable structure of the log-likelihood function
in the distributed setting. 
Upon computing ${\bm u}_{(1)}^t, {\bm v}_{(1)}^t$, we can estimate the top singular value
by $({\bm v}_{(1)}^t)^\top \grd_{\Prm} {\cal L} ( \alp^t, \Prm^{t-1} ) ~{\bm u}_{(1)}^t$
which can also be computed distributively using similar scheme as in the above.
Consequently, the update direction $(\hat{\Prm}^t, \hat{R}^t)$ can be computed 
at the central server using
\beq \label{eq:dt_in}
(\hat{\Prm}^t, \hat{R}^t) = \begin{cases}
({\bm 0}, 0), & \text{if}~\lambda_L \geq ({\bm v}_{(1)}^t)^\top \grd_{\Prm} {\cal L} ( \alp^t, \Prm^{t-1} ) ~{\bm u}_{(1)}^t \eqs,  \\
(-R_{\sf UB}^t {\bm u}_{(1)}^t ({\bm v}_{(1)}^t)^\top, R_{\sf UB}^t), & \text{if}~\lambda_L < ({\bm v}_{(1)}^t)^\top \grd_{\Prm} {\cal L} ( \alp^t, \Prm^{t-1} ) ~{\bm u}_{(1)}^t \eqs.
\end{cases}
\eeq
Lastly, to compute the step size $\beta_t$ required in line~6, an efficient way is to observe
the following decomposition of the inner product:
\beq
\langle \Prm^{t-1} - \hat{\Prm}^t, \grd_{\Prm} {\cal L} ( \alp^t, \Prm^{t-1} ) \rangle
= \sum_{k=1}^K \langle {\cal P}_{\Omega_k} ( \Prm^{t-1} - \hat{\Prm}^t ), \grd_{\Prm} {\cal L}_k ( \alp^t, {\cal P}_{\Omega_k} ( \Prm^{t-1} ) ) \rangle \eqs.
\eeq
This implies that the inner product on the left hand side can be computed 
by aggregating the $K$ terms on the right hand side, where each of 
the $K$ terms can be computed
at the $k$th worker once ${\cal P}_{\Omega_k} ( \hat{\Prm}^t )$ is available. 
As such, the central server also sends ${\cal P}_{\Omega_k} ( \hat{\Prm}^t )$ 
to the workers after \eqref{eq:dt_in}. 
Consequently, the step size is given by:
\beq
\beta_t = \min \Big\{ 1, \frac{ (\widehat{g}_{\Prm} ( \alp^t, \Prm^{t-1}, R^{t-1}; R_{\sf UB}^t ) )_+ }{\sigma_{\Prm} \| {\cal P}_\Omega ( \hat{\Prm}^t - \Prm^{t-1} ) \|_F^2 } \Big\} \eqs,
\eeq
where
\beq
\widehat{g}_{\Prm} ( \alp^t, \Prm^{t-1}, R^{t-1}; R_{\sf UB}^t ) \eqdef 
\langle \Prm^{t-1} - \hat{\Prm}^t, \grd_{\Prm} {\cal L} ( \alp^t, \Prm^{t-1} ) \rangle
+ \lambda_L \big( R^{t-1} - \hat{R}^t \big) \eqs.
\eeq
Note that unlike the function ${g}_{\Prm} (\cdot)$ defined in \eqref{eq:g_fct_real},
the function $\widehat{g}_{\Prm} (\cdot)$ can be negative since the 
matrix $\hat{\Prm}^t$ herein is computed from an inexact pair of top singular vectors.

Several remarks are in order. 
Throughout the optimization, the central server is unaware of the local gradient matrix 
\wrt $\Prm$, instead only its corresponding matrix-vector products 
are transmitted from the workers to the server. In this way, the privacy-sensitive 
data from the workers will not be revealed to the server. 

For any $\delta > 0$, it is well known that in 
high probability (with respect to the random initialization), 
the power method in Algorithm~\ref{alg:dpm} converges \citep{golub2012matrix} 
to an $\delta$-accurate top SVD solution 
in $P = {\cal O}( \log ( 1 / \delta ) )$ steps\footnote{For example, an $\delta$-accurate top SVD solution satisfies 
\beq
\left| ({\bm u}_{(1)}^t)^\top \grd_{\Prm} {\cal L} ( \alp^t, \Prm^{t-1} ) {\bm v}_{(1)}^t -  \sigma_1(\grd_{\Prm} {\cal L} ( \alp^t, \Prm^{t-1} )) \right| \leq \delta \eqs.
\eeq 
In the complexity measure, 
we have hidden the dependency on the spectral gap $\Delta \eqdef \sigma_2(\grd_{\Prm} {\cal L} ( \alp^t, \Prm^{t-1} )) / \sigma_1( \grd_{\Prm} {\cal L} ( \alp^t, \Prm^{t-1} ) ) \leq 1$ in the big-O notation.}. 
Therefore, for the distributed MCGD method, 
the overall complexity required per iteration 
is ${\cal O}( | \Xi | + \max\{n,p\} \log (1 / \delta) )$ at the central server,
and it is ${\cal O}( | \Omega_k | \max\{n,p\} \log (1 / \delta) )$ for the $k$th worker. 
The overall complexity is lower than a centralized implementation especially
when $|\Omega_k| \ll |\Omega|$, e.g., when the number of workers increases.

